%% file: neurips_2026.tex
\documentclass{article}

\usepackage[main,final]{neurips_2026}

\usepackage[utf8]{inputenc} 
\usepackage[T1]{fontenc}    
\usepackage{hyperref}       
\usepackage{url}            
\usepackage{booktabs}       
\usepackage{amsfonts}       
\usepackage{nicefrac}       
\usepackage{microtype}      
\usepackage{xcolor}         

\usepackage{siunitx}
\usepackage{url}
\usepackage{booktabs}       
\usepackage{amsfonts}       
\usepackage{nicefrac}       
\usepackage{microtype}      
\usepackage{xcolor}         
\usepackage{microtype}
\usepackage{graphicx}
\usepackage{booktabs} 
\usepackage{array}
\usepackage{amsmath}
\usepackage{amssymb}
\usepackage{tabularx}
\usepackage{multirow}
\usepackage{wrapfig}
\usepackage{float}
\usepackage{titletoc}
\usepackage{minitoc}
\usepackage{changepage}
\usepackage[normalem]{ulem}
\usepackage{color}
\newcommand{\revision}[1]{#1}

\usepackage{duckuments}         
\usepackage{thmtools,thm-restate}
\usepackage{enumitem}
\usepackage{todonotes}
\usepackage{xcolor,colortbl}
\usepackage{tikz}
\usepackage{tikz-cd}
\usepackage{caption}
\usepackage{subcaption}

\usepackage{lipsum}

\usepackage{amsmath}
\usepackage{amssymb}
\usepackage{mathtools}
\usepackage{amsthm}

\theoremstyle{plain}
\newtheorem{theorem}{Theorem}[section]

\theoremstyle{definition}

\newtheorem{claim}[theorem]{Claim}
\theoremstyle{remark}

\title{Equivariant Spherical Transformer for Efficient Molecular Modeling}

\author{%
  Junyi An\thanks{equal contribution. SAIS: Shanghai Academy of AI for Science} \\
  SAIS \\ Golab \\
  \texttt{anjunyi@sais.com.cn} \\
  \And
  Xinyu Lu$^{*}$\thanks{This work is done when they are interns at SAIS} \\
  Shanghai Innovation Institute \\ Xiamen University \\
  \texttt{xinyulu@stu.xmu.edu.cn} \\
  \And
  Yunfei Shi \\
  SAIS \\ Golab \\
  \texttt{shiyunfei@sais.com.cn} \\
  \And
  Peijia Lin$\dagger$ \\
  Sun Yat-sen University \\ Shanghai Innovation Institute \\
  \texttt{linpj6@mail2.sysu.edu.cn} \\
  \And
  Licheng Xu \\
  SAIS \\ Golab \\
  \texttt{xulicheng@sais.com.cn} \\
  \And
  Qianwei Tang \\
  Nanjing University \\
  \texttt{cleveland225@163.com} \\
  \And
  Chao Qu \thanks{corresponding authors} \\
  Fudan University \\
  \texttt{quchao@fudan.edu.cn} \\
  \And
  Fenglei Cao $\ddagger$ \\
  SAIS, Golab \\ Fudan institute of future technology \\
  \texttt{caofenglei@sais.com.cn} \\
  \And
  Yuan Qi$\ddagger$ \\
  Fudan University \\ SAIS \\
  \texttt{qiyuan@fudan.edu.cn} \\
}

\begin{document}

\maketitle

\begin{abstract}
Equivariant Graph Neural Networks (GNNs) have significantly advanced the modeling of 3D molecular structure by leveraging group representations. However, message passing built only from exactly equivariant operations can have restricted practical expressiveness at a fixed representation degree. We introduce the Equivariant Spherical Transformer (EST), a plug-and-play module that applies Transformer attention to spherical spatial representations obtained from group representations. EST explicitly mixes information across sampled orientations, unlike prior orientation-wise spherical operations, while retaining a strong approximate equivariant inductive bias through near-uniform spherical sampling.
As demonstrated by our experiments on challenging benchmarks like OC20, MPtrj and QM9, EST-based models achieve state-of-the-art performance. For the complex molecular systems within OC20, small models empowered by EST can outperform some larger models and those using additional data. 
We characterize the ideal closed-domain condition under which the continuous operation is exactly equivariant and empirically quantify the finite-sampling accuracy--equivariance--memory trade-off used in practice. \\
\textbf{Code: }https://github.com/ajy112/EST
\end{abstract}

\input{chapter/introduction}
\input{chapter/preliminaries}
\input{chapter/model}

\input{chapter/experiments}
\input{chapter/conclusion}

\section*{Acknowledgments}
We thank our collaborators for their valuable discussions and support. We also thank SAIS and the Golab platform for providing the research environment and infrastructure that supported this work.

\normalem
\bibliographystyle{unsrt}
\bibliography{neurips_2026}


\appendix
\input{chapter/appendix}



\end{document}

%% file: chapter/introduction.tex
\section{Introduction}
Graph Neural Networks (GNNs) have become indispensable for modeling molecular systems and approximating quantum mechanical calculations, providing critical support for high-impact tasks such as drug discovery \citep{senior2020improved,kovacs2025mace,deng2023chgnet} and material design \citep{zitnick2020introduction}. Compared to traditional methods such as Density Functional Theory (DFT), GNNs can predict quantum properties in fractions of a second, significantly reducing computational cost from hours or days. \revision{Among the various GNN architectures, SE(3)-equivariant GNNs \citep{thomas2018tensor,brandstetter2021geometric} are a leading family whose symmetry-aware inductive bias helps capture intricate atomic interactions.}

\begin{figure}[t]
\vskip -0.45in
\begin{center}
\includegraphics[scale=0.40, trim=0cm 0cm 0cm 0cm,clip]{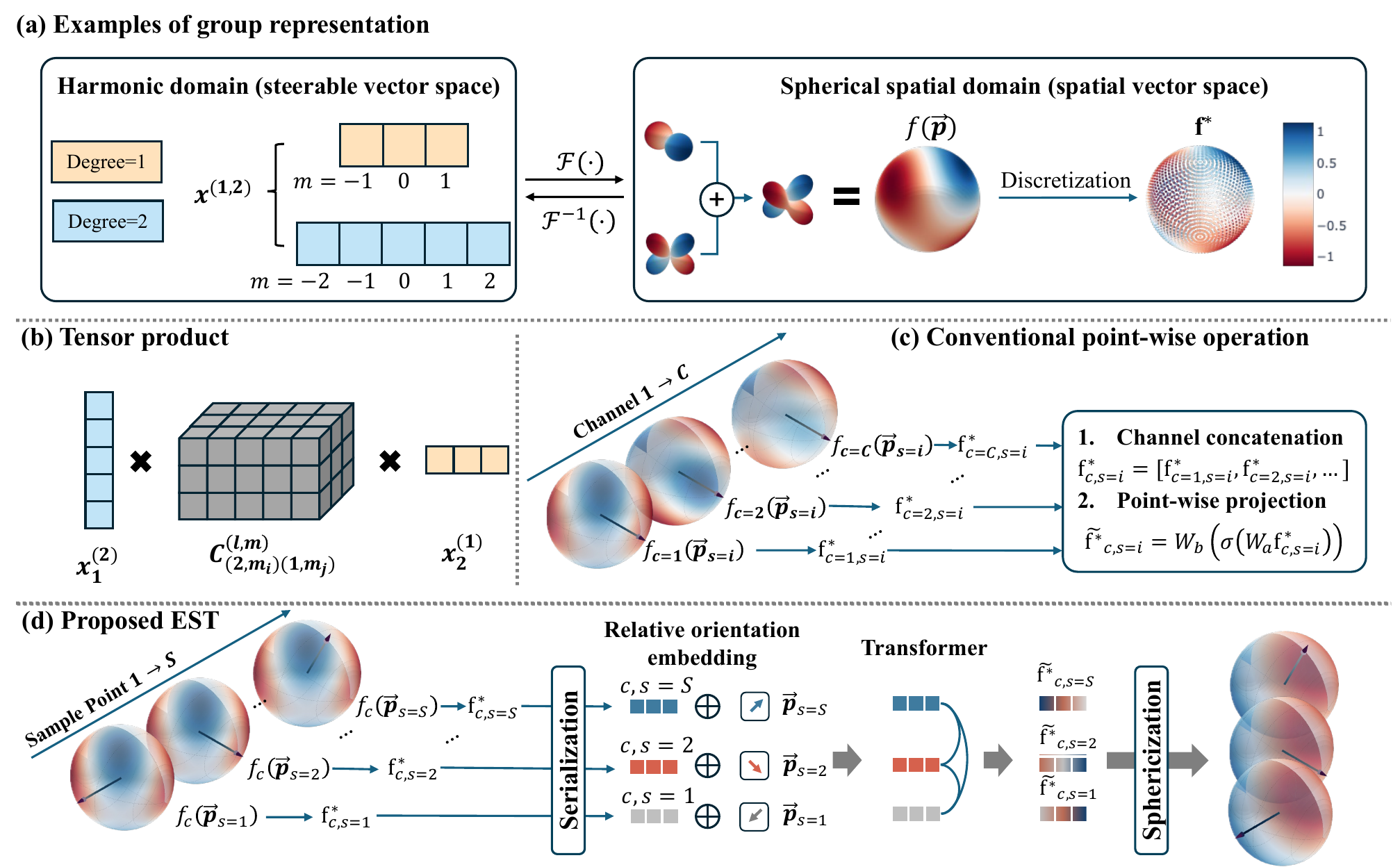}
\caption{
\textbf{Overview of operations in the spherical spatial domain.} (a) \textbf{Examples of group representation.} A Fourier transform is applied to project steerable vectors onto the spatial domain, which are then stored through sampling (i.e., discretization). (b) \textbf{Tensor product operation.} Steerable vectors are combined using the Clebsch-Gordan coefficient $\mathcal{C}$. (c) \textbf{Conventional orientation-wise (point-wise) operation.} Features are processed independently at each sampling orientation $s$. A shared neural network (with linear projection $W$ and activation $\sigma(\cdot)$) acts only across channels ($c=1, \dots, C$) within the same orientation $\vec{\mathbf{p}}_{s=i}$, neglecting cross-orientational relationships. (d) \textbf{Proposed EST operation.} A Transformer-based architecture that explicitly models dependencies across all pairs of orientations ($s=1, \dots, S$), enabling the capture of complex, cross-orientational geometric correlations.
}
\label{fig: grid}
\end{center}
\vskip -0.15in
\end{figure}
In molecular systems, SE(3)-invariance and SE(3)-equivariance are fundamental constraints. For instance, molecular energies remain constant under rotation, while atomic forces rotate in concert with the molecule equivariantly. Adherent to these priors, early invariant models such as SchNet \citep{schutt2018schnet} and HIP-NN \citep{lubbers2018hierarchical} utilize scalar interatomic distances in their message-passing blocks, consequently limiting their ability to capture interactions involving triplets or quadruplets of atoms \citep{miller2020relevance}. Directional GNNs \citep{gasteiger2020directional,gasteiger2021gemnet} address this problem by explicitly incorporating bond angles and dihedral angles. While these models showed performance improvements, their expressiveness remains constrained due to a heavy reliance on handcrafted features. To move beyond these manual descriptors and capture richer geometric information, 
SE(3)-equivariant GNNs employ group representations as node embeddings and construct steerable message-passing blocks \citep{thomas2018tensor,brandstetter2021geometric,liao2023equiformer,aykentgotennet,li2025e2former}, where the tensor product is applied to capture equivariant interactions between group embeddings (see Figure \ref{fig: grid}(b)).
Nevertheless, tensor-product operations provide limited nonlinear mixing at a fixed maximum degree \citep{thomas2018tensor,dym2021on,joshi2023expressive,cen2024high}. Cartesian-tensor models organize higher-degree interactions through symmetric Cartesian tensors and irreducible decomposition \citep{wang2024n}; they retain exact equivariance but their cost grows with correlation order and angular degree. A complementary line, including SCN and eSCN \citep{zitnick2022spherical,passaro2023reducing}, applies an FT to obtain sampled spherical channels and processes each orientation independently. EquiformerV2 adopts related S2Activation and SO(2)-convolution components \citep{liao2024equiformerv}. EST shares this spherical representation but changes the central operation: self-attention mixes all sampled orientations and can model dependencies that an orientation-wise MLP cannot (Figure \ref{fig: grid}(c--d)).

\revision{Finite sampling makes EST approximately, rather than exactly, equivariant for arbitrary rotations. This is a deliberate and measured trade-off. Prior work has likewise studied soft, partial, or approximate equivariance when exact constraints limit expressiveness or the symmetry is imperfect \citep{finzi2021residual,van2022relaxing,wang2022approximately,romero2022learning}. Our goal is to combine an expressive spatial branch with exact-equivariant operators while disclosing the resulting error and memory cost.}

\revision{We present EST as a plug-and-play, approximately equivariant module for high-fidelity modeling of atomic interactions. EST converts Fourier spatial representations into orientation sequences and applies a Transformer to capture dependencies across orientations (Figure \ref{fig: grid}(d)). We use a Fibonacci lattice (FL), optionally refined by repulsion, to reduce finite-sampling error. A controlled replacement of S2Activation in the original SCN isolates the contribution of cross-orientation attention. Across OC20, MPtrj, and QM9, EST improves predictive performance most clearly on complex systems with remaining modeling headroom.}

\revision{\textbf{Our contributions are:} (i) EST introduces cross-orientation attention into spherical-channel models; (ii) we state the exact closed-domain equivariance condition and measure the approximation induced by finite FL sampling; and (iii) controlled ablations and benchmark experiments characterize the resulting accuracy, expressiveness, equivariance, latency, and memory trade-offs.}

%% file: chapter/preliminaries.tex
\section{Preliminaries and Related Work}\label{section:Preliminary}

In this section, we discuss related works and the corresponding mathematical background relevant to equivariant GNNs. 

\paragraph{Notations} We begin by listing the notation commonly used throughout the paper. \textbf{We denote the unit sphere as $\mathbb{S}^{2}$, where the coordinate of a spherical orientation (i.e. spherical sampling point) $\vec{\mathbf{p}} = (\theta, \varphi)$ is represented by its polar angle $\theta$ and azimuth angle $\varphi$.} The symbol $\mathbb{R}$ represents the set of real numbers, while $\mathbf{R}$ denotes a rotation matrix for 3D vectors. We use $[\cdot,\cdot]$ or $\oplus$ to indicate the concatenation of the tensors and $\tilde{\cdot}$ to represent the update of the tensors, respectively.

\paragraph{Equivariance and Invariance} Given a group $\mathbf{G}$ and a transformation parameter $g \in \mathbf{G}$, a function $\phi: \mathcal{X} \to \mathcal{Y}$ is said to be equivariant to $g$ if it satisfies:
\begin{equation}
    \label{equ:equivar}
     \phi(T(g)[x]) = T'(g)[\phi(x)],
\end{equation}
where $T'(g): \mathcal{Y} \to \mathcal{Y}$ and $T(g):\mathcal{X} \to \mathcal{X}$ denote the corresponding transformations on $\mathcal{Y}$ and $\mathcal{X}$, respectively. Invariance is a special case of equivariance where $T'(g)$ is the identity transformation. 

\paragraph{Spherical Harmonics and Steerable Vectors} Spherical harmonics, a set of orthonormal basis functions defined over the sphere $\mathbb{S}^2$, are commonly employed in equivariant models. The real-valued spherical harmonics are typically denoted as $\{ Y^{(l,m)}: \mathbb{S}^{2} \to \mathbb{R} \}$, where $l$ represents the degree and $m$ represents the order. For any sample point $\vec{\mathbf{p}}$, we define $\mathbf{Y}^{(l)}(\vec{\mathbf{p}}) = [ Y^{(l,-l)}(\vec{\mathbf{p}}), Y^{(l,-l+1)}(\vec{\mathbf{p}}),..., Y^{(l,l)}(\vec{\mathbf{p}}) ]$, a vector of size $2l+1$, and $\mathbf{Y}^{(0\to l)}(\vec{\mathbf{p}}) = [\mathbf{Y}^{(0)}(\vec{\mathbf{p}}), ..., \mathbf{Y}^{(l)}(\vec{\mathbf{p}})]$, a vector of size $(l+1)^2$.

A key property of spherical harmonics is their behavior under rotation $\mathbf{R} \in SO(3)$:
\begin{equation}
\label{equ:Wigner-D}
    \mathbf{Y}^{(l)}(\mathbf{R}\vec{\mathbf{r}}) = \mathbf{D}^{(l)}(\mathbf{R}) \mathbf{Y}^{(l)}(\vec{\mathbf{r}}),
\end{equation}
where $\mathbf{D}^{(l)}(\mathbf{R})$ is a $(2l+1) \times (2l+1)$ matrix known as the Wigner-D matrix of degree $l$. Notably, $\mathbf{D}^{(1)}(\mathbf{R}) = \mathbf{R}$. Thus, $\mathbf{R}$ and $\mathbf{D}^{(l)}(\mathbf{R})$ can represent $T(g)$ and $T'(g)$ in equation \eqref{equ:equivar}. Following the convention in \citep{chami2019hyperbolic,brandstetter2021geometric}, we say that $\mathbf{Y}^{(l)}(\vec{\mathbf{p}})$ is steerable by the Wigner-D matrix of the same degree $l$. Furthermore, a vector that transforms according to an $l$-degree Wigner-D matrix is termed an $l$-degree steerable vector or a type-$l$ vector, residing in the vector space $\mathbb{V}_{l}$. Further mathematical details are available in Appendix \ref{app:spherical_harmonic_detail}.

\paragraph{Equivariant Architectures} Most equivariant models are based on the message-passing neural network (MPNN), the details of which are provided in the Appendix \ref{app:mpnn}. A common practice in equivariant models is to use steerable vectors as node embedding and encode relative positions $\vec{\mathbf{r}}_{ij}$ using spherical harmonics as edge attributes. TFN \citep{thomas2018tensor} and NequIP \citep{batzner2022} provide a general framework that learns the interaction between node embeddings and edge attributes through equivariant convolution filters. These filters are composed of equivariant operations such as degree-wise linear layers, Clebsch-Gordan (CG) tensor products, and Gate mechanisms. SEGNN \citep{brandstetter2021geometric} extends this convolution framework by more non-linear operations to achieve enhanced learning ability. SE(3)-Transformer \citep{fuchs2020se} and Equiformer \citep{liao2023equiformer} introduce an attention mechanism that assigns rotation-invariant weights to neighbor attributes, thereby improving the learning of key atomic interactions. 
Several studies \citep{passaro2023reducing,luo2024enabling,li2025e2former} have sought to develop variants of the tensor product operation to reduce computational complexity without significantly compromising expressiveness. Nevertheless, the core operation in these works remains the convolution filter based on the CG tensor product. In contrast, GotenNet \citep{aykentgotennet} proposes an effective tensor-product-free structure that combines neighbor attention with hierarchical refinement mechanisms, achieving SOTA performance in certain molecular benchmarks. Our work tends to explore a distinct tensor-product-free paradigm: we project steerable vectors into the spherical spatial domain, enabling a novel attention mechanism to weight embedding compositions directly. This approach fundamentally diverges from previous attention-based GNNs by shifting the focus from local neighbor-indexing to spatial-domain interactions, providing a more direct way to capture geometric dependencies.


\paragraph{Spherical Fourier Transform} Any square-integrable function $f(\vec{\mathbf{p}})$ defined over the sphere $\mathbb{S}^2$ can be expressed in a spherical harmonic basis via the spherical Fourier transform $\mathcal{F}$:
\begin{equation}
\label{equ:fft}
    f(\vec{\mathbf{p}}) = \mathcal{F}(\mathbf{x}) = \sum_{l=0}^{\infty}\sum_{m=-l}^{l}x^{(l,m)}Y^{(l,m)}(\vec{\mathbf{p}}),
\end{equation}
where $x^{(l,m)}$ is the spherical Fourier coefficient. Conversely, the coefficient can be obtained from the function $f(\vec{\mathbf{p}})$ via the inverse transform $\mathcal{F}^{-1}$:
\begin{equation}
\label{equ:ifft}
    x^{(l,m)} = \mathcal{F}^{-1}(f(\vec{\mathbf{p}})) = \int_{\Omega} f(\vec{\mathbf{p}})Y^{(l,m)*}(\vec{\mathbf{p}})d\Omega,
\end{equation}
where $d\Omega$ denotes the unit sphere and $Y^{(l,m)*}(\cdot)$ is the conjugate of the spherical harmonic function. The function $f(\vec{\mathbf{p}})$ and its coefficients $\mathbf{x}$ are referred to as the representations in the spherical spatial domain and the harmonic domain, respectively. These processes are illustrated in Figure \ref{fig: grid}(a). SCN \citep{zitnick2022spherical} introduced an approach that applies a simple point-wise neural network in the spherical spatial domain. This technique has been subsequently adopted by several works \citep{passaro2023reducing,liao2024equiformerv,fu2025learning} within their message-passing frameworks. However, point-wise operations inherently limit expressiveness as they process each point in isolation, failing to account for the relationships and structural correlations between different points in the spatial domain. In this work, we empower models to learn across the entire spherical spatial domain, thereby capturing complex cross-orientational dependencies.

%% file: chapter/model.tex
\section{Model}\label{sec:model}
In this section, we first make a review of the tensor product to detail our motivation. Then, we introduce the formulation of our EST and clarify its properties.

\subsection{Review of traditional equivariant operations}\label{sec:CGTP}
A common wisdom in equivariant models is to incorporate steerable vectors with multiple degrees for node or edge features and then employ CG tensor products $\otimes: \mathbb{V}_{l_1} \times \mathbb{V}_{l_2} \to \mathbb{V}_l$, which are defined as:
\begin{scriptsize}
\begin{equation}
\label{equ:CG}
 (\mathbf{x}_1 \otimes \mathbf{x}_2)^{(l,m)} = \sum^{l_{1}}_{m_{1}=-l_{1}} \sum^{l_{2}}_{m_{2}=-l_{2}} \mathcal{C}^{(l,m)}_{(l_{1},m_{1})(l_{2},m_{2})} \mathbf{x}^{(l_{1}, m_{1})}_1\mathbf{x}^{(l_{2}, m_{2})}_2,
\end{equation}
\end{scriptsize}
where $\mathcal{C}^{(l,m)}_{(l_{1},m_{1})(l_{2},m_{2})}$ are the CG coefficients, a sparse tensor yielding non-zero terms when $|l_{1}-l_{2}| \le l \le (l_{1} + l_{2})$. These products enable interactions between various combinations of $l_1$ and $l_2$, crucial for the universal expressiveness of MPNNs \citep{dym2021on,joshi2023expressive}. However, CG tensor products provide limited non-linearity \citep{brandstetter2021geometric}, often requiring the stacking of multiple layers to capture complex molecular features. Furthermore, their high complexity makes it difficult to apply to learning large-scale cross-channel interactions \citep{liao2023equiformer}, especially in message block. While techniques like Gate mechanisms \citep{brandstetter2021geometric,liao2023equiformer} and geometry-aware tensor attention \citep{aykentgotennet} can be used to mitigate this limitation, they apply non-linear transformations to the invariant ($l = 0$) representations. Higher-degree representations ($l>0$) are then updated through a unified scaling multiplication:
\begin{equation}
\label{equ:previousgate}
C_{inv} = \mathrm{NL}(\mathbf{x}^{(0)}, \sum^{L}_{l=1}\mathrm{d}(\mathbf{x}^{(l)})), \quad \text{and} \quad \tilde{\mathbf{x}}^{(l)} = C_{inv} \cdot \mathbf{x}^{(l)},
\end{equation}
where $C_{inv}$ is an invariant scalar, $\mathrm{NL}(\cdot)$ is a non-linear network, and $\mathrm{d}(\cdot): \mathbb{V}_{l} \rightarrow \mathbb{V}_{0}$ maps equivariant features to invariants, such as an inner product. This indirect exchange of information between higher-degree ($l>0$) representations still limits the overall expressiveness.

\subsection{Equivariant Spherical Transformer}\label{sec:est}
\subsubsection{Node embedding}

\paragraph{Steerable Representation} For representation with steerable vectors, we first define a maximal degree $L$. Each node $n$'s embedding $\mathbf{x}_{n}$ comprises $C$ channels, and each channel $c$ is a concatenation of steerable vectors from degree $0$ to $L$: $\mathbf{x}^{(0\to L)}_{n,c} = [x^{(0)}_{n,c},\mathbf{x}^{(1)}_{n,c},...,\mathbf{x}^{(L)}_{n,c}]$. As a result, the dimension of a steerable node embedding is $(L+1)^2 \times C$. 

\paragraph{Spatial Representation} Steerable representations $\mathbf{x}_{n,c}$ can be transformed into a spherical function $f_{n,c}(\vec{\mathbf{p}})$ via equation \eqref{equ:fft}, with the summation truncated at a maximum degree $L$. Furthermore, $f_{n,c}(\vec{\mathbf{p}})$ is discretely represented by sampling $S$ points on the sphere, yielding a spatial node embedding $\mathbf{f}^{*}_{n} \in \mathbb{R}^{S \times C}$. It can be expressed as the concatenation over sampled points:
\begin{equation}
\label{equ:discretefft}
    \mathbf{f}^{*}_{n,c} = \Big[ (\mathbf{x}_{n,c}^{(0\to L)})^T \mathbf{Y}^{(0\to L)}(\vec{\mathbf{p}}_1), \, (\mathbf{x}_{n,c}^{(0\to L)})^T \mathbf{Y}^{(0\to L)}(\vec{\mathbf{p}}_2), \dots, \, (\mathbf{x}_{n,c}^{(0\to L)})^T \mathbf{Y}^{(0\to L)}(\vec{\mathbf{p}}_S) \Big].
\end{equation}
We use the term $\mathbf{f}^{*}_{n,c,s}$ to denote the signal at the sampled point $\vec{\mathbf{p}}_s$ of channel $c$ of node $n$. The inverse transform in equation \eqref{equ:ifft} can convert $\mathbf{f}^{*}_{n}$ back to $\mathbf{x}_{n}$. 


\paragraph{The Significance of Operating in the Spatial Domain} In the spatial representation, the explicit degree and order are no longer directly accessible. Instead, higher-degree information is encoded as geometric features distributed across the spherical surface, as illustrated in Figure \ref{fig: grid}(a). Intuitively, \textbf{by modeling the interdependencies of these geometric features, the model can effectively exchange information between implicit degrees, thereby approximating the capability of the tensor product.} \revision{SCN/eSCN-style S2Activation processes channels independently at each orientation \citep{zitnick2022spherical,passaro2023reducing}; EST instead models dependencies among all orientations. We analyze two distinct properties: exact equivariance for an ideal rotationally closed domain and approximate equivariance after finite sampling, together with evidence that cross-orientation attention increases practical expressiveness.}

\subsubsection{Spherical Attention}\label{sec:ssa}
To prepare the data, we flatten all sampled points from $\mathbf{f}^{*}$ into a sequential format, where each point $s$ represented by a $C$-dimensional feature vector. We then apply an attention mechanism \citep{vaswani2017attention}:
\begin{equation}
\label{equ:attention_est}
    a_{s=i, s=j} = \frac{\exp(\mathbf{Q}_{s=i} \mathbf{K}_{s=j}^T / \sqrt{C})}{\sum_{k=1}^{S} \exp(\mathbf{Q}_{s=i} \mathbf{K}_{s=k}^T / \sqrt{C})},
    \text{and} \quad \tilde{\mathbf{f}}^{*}_{s=i} = \sum_{j=1}^{S} a_{s=i, s=j} \mathbf{V}_{s=j}.
\end{equation}
where the $\mathbf{Q}$, $\mathbf{K}$, and $\mathbf{V}$ matrices are obtained through point-wise linear transformations of $\mathbf{f}^{*}_{s=i}$. The spherical attention is the core of EST. We also incorporate the pre-LayerNorm \citep{xiong2020layer} and feedforward networks (FFN) in EST. As a fundamental building block, EST focuses on interactions within an equivariant representation while preserving input-output dimensionality. 




\paragraph{Equivariance and sampling strategies} \revision{We first state the exact result on the closed domain required by the proof, and then distinguish it from the finite sampling used in implementations.}

\begin{theorem}\label{theorem:equivariance}
\revision{Let $\Omega\subseteq\mathbb{S}^{2}$ be the domain of the spherical transform, equipped with the rotation-invariant surface measure, and assume that it is closed under every $\mathbf{R}\in SO(3)$, i.e., $\mathbf{R}\Omega=\Omega$. If spherical attention is evaluated over $\Omega$ before the inverse transform, then the continuous EST operator is $SO(3)$-equivariant:}
\begin{equation}
\mathcal{F}^{-1}\big(\mathrm{EST}\big(\mathcal{F}(\mathbf{D}\mathbf{x})\big)\big) = \mathbf{D}\mathcal{F}^{-1}\big(\mathrm{EST}\big(\mathcal{F}(\mathbf{x})\big)\big),
\end{equation}
\revision{where $\mathbf{D}$ is the Wigner-D representation associated with $\mathbf{R}$.}
\end{theorem}

\revision{The proof is provided in Appendix \ref{app:proof_equivariance}. The theorem applies to the closed continuous domain $\Omega$ (in our setting, the full sphere), not to an arbitrary finite point set. A finite set $\mathcal{P}=\{\vec{\mathbf p}_s\}_{s=1}^{S}$ is closed only under its grid-symmetry subgroup, so arbitrary rotations introduce discretization error. We therefore seek near-uniform samples that make the quadrature and nearest rotated-point mismatch small. Compared with the latitude-dependent e3nn grid \citep{e3nn}, we use Fibonacci Lattice (FL) sampling \citep{gonzalez2010measurement}:}
\begin{equation}
\label{equ:FL_sampling}
    \vec{\mathbf{p}}_s = [\sqrt{1-z_s^2}\cos(2s\pi/\lambda), \sqrt{1-z_s^2}\sin(2s\pi/\lambda), z_s],
\end{equation}
where $z_s = \frac{2s-1}{S}-1$ and $\lambda = \frac{1+\sqrt{5}}{2}$ denotes the golden ratio. Equation \eqref{equ:FL_sampling} ensures that the sampled points are distributed across the spherical surface with approximately uniform density (see Figure \ref{fig:sampling}(b)). While FL sampling provides high uniformity, we further refine this distribution by simulating point motion on the sphere via consistent repulsions. This process resembles a simple molecular dynamics simulation where each "atom-like" point is confined to the spherical surface, with repulsive forces determined by inter-point distances. Following FL sampling, we define the self-conjugate basis for the FT and its inverse. Further details regarding the sampling methods and the basis are provided in Appendix \ref{app:proof_equivariance}. Furthermore, our ablation experiments (Table \ref{tab:equivariance}) demonstrate that sampling methods with poor uniformity can severely compromise equivariance.

\paragraph{Expressiveness} 
As demonstrated by \citep{dym2021on,joshi2023expressive,cen2024high}, the expressiveness of CG tensor product-based modules is constrained by the maximum degree $L$, a finding corroborated by ablation experiments. In contrast, our proposed EST module can project steerable representations of any degree onto the sphere and perform flexible, learnable operations. This allows the resulting features to be reprojected back to any desired degree, thereby potentially overcoming the expressiveness limitations.
\begin{claim}[Unproven expressiveness intuition]\label{theorem:expressiveness}
\revision{Under information-preserving spherical transforms, cross-orientation Transformer mixing may approximate the local functions represented by CG tensor products and also represent functions outside the corresponding fixed-degree local class.}
\end{claim}
\revision{Claim \ref{theorem:expressiveness} is not a proved set-inclusion theorem. It is intuition supported by the universal-approximation argument in Appendix \ref{app:proof_expressiveness}, the controlled fitting experiment in Appendix \ref{app:controlled_ablation}, and the symmetry task in Table \ref{tab:rotsym}. The transform itself is information-preserving once the sampling meets the relevant band-limit/Nyquist requirement (we use $S\ge (2L)^2$) \citep{zitnick2022spherical}.}

\paragraph{Relative Orientation Embedding} 
In basic form, spherical attention determines the dependencies between points based solely on their features, ignoring their relative positions in the input sequence. This means the output feature of a given point remains unchanged even if the order of all other points is altered (Figure \ref{fig:position_embedding}). Such behavior is undesirable, as different point orderings correspond to distinct spherical functions. To enable the model to distinguish between these configurations, we introduce a relative orientation embedding by augmenting the query ($\mathbf{Q}$) and key ($\mathbf{K}$) vectors with point information:
\begin{equation}
\label{equ:position_embed}
    \mathbf{Q}_{s=i} := [\mathbf{Q}_{s=i}, \vec{\mathbf{p}}_{s=i}], \quad \mathbf{K}_{s=j} := [\mathbf{K}_{s=j}, \vec{\mathbf{p}}_{s=j}],
\quad \forall i,j \in \{1, \dots, S\}.
\end{equation}
Consequently, the inner product between $\mathbf{Q}_{s=i}$ and $\mathbf{K}_{s=j}$ now incorporates a term related to the orientation-aware inner product $\vec{\mathbf{p}}^{T}_{s_i}\vec{\mathbf{p}}_{s_j}$. Crucially, this augmentation does not compromise the equivariant inductive bias of EST, as proven in Appendix \ref{app:proof_relativeembedding}. Our approach shares conceptual similarities with Rotary Position Embeddings (RoPE) \citep{su2024roformer} in Natural Language Processing (NLP), where the relative position embedding is invariant under transformation. However, our method fundamentally differs from RoPE by addressing rotation invariance instead of permutation invariance.

\subsubsection{Integrating EST into Equivariant Architectures}\label{sec:MoE_architecture}
The modular design of EST enables seamless integration into existing equivariant neural networks, circumventing complex hyperparameter tuning while leveraging the inherent strengths of diverse architectures. \textbf{Given EST’s high expressivity and the proficiency of Transformers in multi-channel processing, we incorporate this framework into the update blocks of various models to handle complex message aggregation.} Furthermore, because most standard update blocks are architecturally minimalist, EST can be easily deployed as a versatile, plug-and-play enhancement.

\revision{In our implementation, the base update and an EST branch run in parallel and are combined by learned invariant weights. We call them steerable and spatial experts only as shorthand; this is not a sparse or routed language-model MoE. The mixed branch compensates for finite-sampling error while adding cross-orientation capacity. Appendix \ref{app:ab_modules} shows that a minimal $1+1$ mixture is more effective than merely adding same-type branches.}

%% file: chapter/experiments.tex
\newcommand{\mcc}[1]{\multicolumn{1}{c}{#1}}
\section{Experiments}\label{sec:experiments}
In this section, we construct experiments to investigate the efficacy of the proposed EST. We evaluate its performance on the OC20 (S2EF and IS2RE), MPtrj and QM9 benchmarks. Our analysis involves a comparison with a wide range of baseline models, including SchNet \citep{schutt2018schnet}, PaiNN \citep{schutt2021equivariant}, SEGNN \citep{brandstetter2021geometric}, TFN \citep{thomas2018tensor}, DimeNet++ \citep{klicpera2020fast}, Equiformer \citep{liao2023equiformer}, EquiformerV2 \citep{liao2024equiformerv} and EScAIP \citep{qu2024importance}. Specifically for the OC20 benchmark, we extend comparison to include SpinConv \citep{shuaibi2021rotation}, GemNet \citep{gasteiger2021gemnet,gasteiger2022gemnetoc}, SphereNet \citep{liu2021spherical}, SCN \citep{zitnick2022spherical} and eSCN \citep{passaro2023reducing}. For the MPtrj benchmark, our comparison includes CHGNet \citep{deng2023chgnet} and MACE \citep{batatia2022mace}. Similarly, our QM9 experiments are augmented with comparisons to TorchMD-NET \citep{tholke2022equivariant}, EQGAT \citep{le2022equivariant} and GotenNet \citep{aykentgotennet}. The baseline configurations can be found in Appendix \ref{app:hp_baseline}.
\begin{table}[tbp]
\vskip -0.1in
\caption{S2EF validation results. $\lambda_{E}$ is the coefficient of the energy loss. ``$^\tau$'' denotes the test results.}
    \label{tab:oc20_s2ef}
    \centering
    \sisetup{round-mode=figures, round-precision=3}
    \resizebox{1\linewidth}{!}{
    \begin{tabular}{c l rr  S[table-format=3] S[table-format=2.1] }
    \toprule
    & & \mcc{Number of} & Throughput & \mcc{Energy MAE}    & \mcc{Force MAE}          \\
    Dataset & Model & \mcc{parameters} & Samples/s & \mcc{\si{(\milli\electronvolt)} $\downarrow$} & \mcc{\si[per-mode=symbol]{(\milli\electronvolt\per\angstrom)} $\downarrow$}  \\ 
     \midrule[1.2pt]
        \multirow{2}{*}{All+MD} & EScAIP-Small & 83M &  -  &       \textbf{229} &           15.1   \\
        & EquiformerV2 ($\lambda_E=4$, 8 layers) &  31M & 7.1  &  232 &         16.26  \\
        \midrule
        \multirow{10}{*}{All} & SchNet  & 9.1M &      -  &              548.525 &                          56.8 \\
        & DimeNet++-L-F+E  & 10.7M &  4.6  &                       515.475 &                         32.75 \\ 
        & SpinConv  & 8.5M &     6.0  &                 371.01165 &                     41.197875 \\
        & GemNet-dT  & 32M &         25.8  &            315.118775 &                     27.172475 \\
        & GemNet-OC & 39M &    18.3  &      244.28979999999999 &        21.719050000000003 \\
        & SCN$^\tau$ & 271M &  -  &   244 &            17.7  \\
        & eSCN$^\tau$ & 200M &  2.9  &       242 &           17.1   \\
        & EquiformerV2 ($\lambda_E=2$, 20 layers) &  153M & 1.8  &  236 &   15.7   \\
        & EquiformerV2 + EST ($\lambda_E=4$, 8 layers) &  45M & 6.8  &  \textbf{229} &         15.8  \\
        & EquiformerV2 + EST ($\lambda_E=2$, 8 layers) &  45M & 6.8  &  232 &         \textbf{15.0}  \\
        \bottomrule
    \end{tabular}
    }
    \vskip -0.25in
\end{table}
\subsection{OC20 Results}\label{sec:oc20results}
\paragraph{Dataset and Configurations} The OC20 dataset, comprising over 130 million molecular structures, covers a broad spectrum of materials and adsorbates. We evaluate models on two core sub-datasets: S2EF and IS2RE. All configurations are detailed in Appendix \ref{app:hp_est}. Notably, We did not use very deep models. For instance, our S2EF experiments utilize a 8-layer architecture, and IS2RE experiments employ a 6-layer one, significantly fewer than many advanced methods. 

\paragraph{S2EF Results} Using EquiformerV2 as the base model, we implemented an EST-based variant and conducted training on the S2EF-All dataset. Each validation set was divided into four similarly-sized subsets: In-Domain (ID), Out-of-Domain Catalysis (OOD Cat), Out-of-Domain Both (OOD Both). Consistent with prior work \citep{passaro2023reducing,liao2024equiformerv}, the best model was selected based on its performance on the ID validation subset during training and subsequently evaluated across all validation subsets. 
We compared our results with two variants of the EquiformerV2 base model: a deep 20-layer architecture and one trained with additional MD dataset. As shown in Table \ref{tab:oc20_s2ef}, EST outperformed both EquiformerV2 variants in energy and forces prediction. Furthermore, EST achieved competitive results in force prediction, surpassing several bigger baselines including EScAIP, SCN and eSCN.
\begin{table}[tbp]
\caption{IS2RE results of models trained on IS2RE training dataset. Bold and underline indicate the best result, and the second best result, respectively.}\label{tab:IS2REDirect}
\resizebox{\linewidth}{!}{
\begin{tabular}{l|cccc|cccc}
\toprule
                & \multicolumn{4}{c|}{Energy MAE (meV) $\downarrow$}       & \multicolumn{4}{c}{EwT (\%) $\uparrow$}      \\
Model           & ID   & OOD Ads & OOD Cat & OOD Both               & ID   & OOD Ads & OOD Cat & OOD Both \\ \midrule
SchNet          & 639  & 734     & 662     & 704                    & 2.96 & 2.33    & 2.94    & 2.21     \\
PaiNN           & 575  & 783     & 604     & 743                    & 3.46 & 1.97    & 3.46    & 2.28     \\
TFN            & 584  & 766     & 636     & 700                    & 4.32 & 2.51    & 4.55    & 2.66     \\
DimeNet++       & 562  & 725     & 576     & 661                    & 4.25 & 2.07    & 4.10    & 2.41     \\
GemNet-dT       & 527  & 758     & 549     & 702                    & 4.59 & 2.09    & 4.47    & 2.28     \\
GemNet-OC       & 560  & 711     & 576     & 671                    & 4.15 & 2.29    & 3.85    & 2.28     \\
SphereNet       & 563  & 703     & 571     & 638                    & 4.47 & 2.29    & 4.09    & 2.41     \\
SEGNN           & 533  & 692     & 537     & 679                    & \textbf{5.37} & 2.46    & \uline{4.91}    & 2.63     \\
SCN & 516  & \textbf{643}     & 530     & \uline{604}            & 4.92 & \textbf{2.71}   & 4.42    & \textbf{2.76}     \\ 
\midrule
Equiformer      & \uline{504}  & 688     & \uline{521}     & 630    & 5.14 & 2.41    & 4.67    & \uline{2.69}     \\
+ EST     & \textbf{501}  & \uline{652}  & \textbf{502} & \textbf{578} & \uline{5.16} & \uline{2.67}    & \textbf{5.16}    & \textbf{2.76}     \\
\bottomrule
\end{tabular}
}
\vskip -0.3in
\end{table}
\paragraph{IS2RE Results} We trained our model directly on the IS2RE training set for energy prediction. 
Related results, including the Mean Absolute Error (MAE) for energy and the percentage of Energies Within a Threshold (EwT) for each model, are summarized in Table~\ref{tab:IS2REDirect}. We use a 6-layer Equiformer as base model. EST achieves SOTA performance on ID, OOD Cat, OOD Both and the second best on OOD Ads. It is notable that the current SOTA model on OOD Ads, SCN, employs complex 16-layer architecture with over 100M parameters while EST achieves closely competitive performance using a unified 6-layer structure comprising just 32.47 million parameters.
Furthermore, we observe that EST consistently surpasses the base model (Equiformer) across all evaluated metrics, providing further evidence for the effectiveness of the proposed EST module. 
\subsection{MPtrj Results} 
\paragraph{Dataset and Configurations} The MPtrj benchmark consists of a large-scale collection of material molecular dynamics trajectories for predicting energy, force, and stress. We integrate EST into a lightweight 8-layer EquiformerV2. Note that we compare our approach against the EquiformerV2 and DeNS \citep{liao2024generalizing} baselines as reported in \citep{barroso2024open}, the latter being an efficient self-supervised enhancement technique. Other configurations are detailed in Appendix \ref{app:hp_est}.

\paragraph{Results} Note that molecules in MPtrj show complex interaction where metallic systems typically exhibit complex electronic orbital structures and interfacial properties. Remarkablely, From Table \ref{tab:results_mptrj}, we observe that EST significantly outperform other baselines on all the metrics, showing a stronger expressiveness. 
Furthermore, due to discrepancies and ambiguities in the hyperparameter configurations of prior work, we also reproduced the results of the base model in the Appendix \ref{app:ab_modules}, where EST-based model still demonstrates significant improvement. Additional NVE dynamics, force-equivariance, geometry-relaxation, and vibrational tests in Appendix \ref{app:physical_validity} show that EST remains applicable to interatomic-potential workflows, while also revealing the non-conservativity limitation of a direct force head.
\begin{table}[tbp]
\vskip -0.2in
\small
\centering
\caption{Results on Mptrj dataset.}\label{tab:results_mptrj}
\begin{tabular}{l|ccc}
\toprule
Model & Energy (meV/atom) $\downarrow$ & Forces (meV/\AA)  $\downarrow$ & Stress (meV/\AA$^3$)  $\downarrow$ \\
\midrule
CHGNet & 33 & 79  & 2.19 \\
MACE & 20 & 45 & - \\
\midrule
EquiformerV2-S & 12.4 & 32.22 & 1.55 \\
 + DeNS & 11.43 & 31.67 & 1.44 \\
 + EST & \textbf{10.80} & \textbf{28.85} & \textbf{1.36} \\
\bottomrule
\end{tabular}
\vskip -0.2in
\end{table}

\begin{table}[tbp]
\caption{Results on QM9 dataset for various properties. $\dag$ denotes using different data partitions. Gray indicates the optimal value in the comparison between EST and the base model.}\label{tab:QM9}
\resizebox{\linewidth}{!}{
\begin{tabular}{l|cccccccccccc}
\toprule
Task              & $\alpha$     & $\Delta \varepsilon$ & $\varepsilon_{HOMO}$ & $\varepsilon_{LUMO}$ & $\mu$ & $C_{v}$     & $G$ & $H$ & $R^{2}$  & $U$ & $U_{0}$ & $ZPVE$ \\
Units             & $bohr^{3}$     & meV                  & meV                  & meV                  & D     & cal$/$(mol K) & meV & meV & $bohr^{3}$ & meV & meV     & meV  \\ \midrule
SchNet            &.235          &63                    &41 &34 &.033   &.033 &14         &14           &.073 &19   &14   &1.70  \\
TFN$^{\dag}$        & .223        & 58 & 40 & 38 & .064 & .101 & - & - & - & -  &- & - \\
DimeNet++         &.044          &33          &25 &20 &.030   &.023 &8 & 7  & .331         & 6   & 6   & 1.21 \\
PaiNN             & .045        & 46 & 28 & 20 & .012 & .024 & 7.35 & 5.98 & .066 & 5.83 & \uline{5.85} & 1.28 \\
TorchMD-NET       & .059        & 36         &20 &18 &\uline{.011}   &.026 & 7.62      & 6.16        &.033 & 6.38 & 6.15 & 1.84 \\
SEGNN$^{\dag}$      & .060        & 42         &24 &21 &.023   &.031 &15         & 16          & .660    &13    &15      & 1.62 \\
EQGAT    &.053           &32          &20 &16 &.011 &.024 & 23 & 24 &.382 &25 &25 &2.00 \\
EquiformerV2        &.050          &29          &14 &13 & .010 &.023 & 7.57 & 6.22 &.186 & 6.49 & 6.17 & 1.47 \\
\midrule
Equiformer        &.046         &30          &15 &14 &\cellcolor{gray!40}.011 &.023 & 7.63 & 6.63 &.251 & 6.74 & 6.59 & \cellcolor{gray!40}1.26 \\
+ EST     &\cellcolor{gray!40}.041         & \cellcolor{gray!40}29          &\cellcolor{gray!40}14 &\cellcolor{gray!40}13 &\cellcolor{gray!40}.011 &\cellcolor{gray!40}.021 & \cellcolor{gray!40}7.18 & \cellcolor{gray!40}6.17 &\cellcolor{gray!40}.227 & \cellcolor{gray!40}6.35 & \cellcolor{gray!40}6.32 & 1.27 \\
\midrule
GotenNet-S      &.033         &\cellcolor{gray!40}\textbf{21.2}          &16.9 &13.9 &.0075 &.020 & 5.50 & \cellcolor{gray!40}\textbf{3.70} &\cellcolor{gray!40}\textbf{.029} & \cellcolor{gray!40}\textbf{3.67} & 3.71 & \cellcolor{gray!40}\textbf{1.09} \\
+ EST    &\cellcolor{gray!40}\textbf{.030}         &31.2          &\cellcolor{gray!40}\textbf{16.7} &\cellcolor{gray!40}\textbf{13.3} &\cellcolor{gray!40}\textbf{.0070} &\cellcolor{gray!40}\textbf{.019} & \cellcolor{gray!40}\textbf{5.40} & 3.71 &\cellcolor{gray!40}\textbf{.029} & 3.70 & \cellcolor{gray!40}\textbf{3.70} & 1.19 \\
\bottomrule
\end{tabular}
}
\vskip -0.1in
\end{table}
\subsection{QM9 Results}\label{sec:qm9results}
\paragraph{Dataset and Configurations} The QM9 benchmark comprises quantum chemical properties for a relevant, consistent, and comprehensive chemical space of 134k equilibrium small organic molecules containing up to 29 atoms. We developed two model architectures: 
the first model, denoted as Equiformer+EST, integrates Equiformer's message block with EST's update block. The second model, denoted as GotenNet+EST, integrates GotenNet's message block with EST's update block. For additional configuration specifics, please refer to Appendix \ref{app:hp_est}.

\paragraph{Results} \revision{QM9 is small and contains only equilibrium molecules with at most nine heavy atoms, so performance is comparatively saturated and symmetry is a strong regularizer. Equiformer+EST improves ten of twelve tasks, but the gains are attenuated for deeper GotenNet backbones. Our depth ablation is consistent with two contributing effects: less remaining accuracy headroom and accumulation of finite-sampling equivariance error with depth. We report this as a limitation and interpretation rather than a conclusively isolated causal mechanism.}

\subsection{Ablation study}\label{sec:ablation}
In this section, we explore three key properties of EST: expressiveness, equivariance and latency. The other experiments can be found in Appendix \ref{app:ab_modules}, where we supplement theoretical experiments on expressiveness and comprehensively investigate the influence of building components, including self-attention, spherical relative orientation embedding and hybrid experts.

\paragraph{\revision{Controlled replacement}} \revision{In the original SCN on OC20-S2EF-200K, replacing only its orientation-wise grid MLP with EST improves force MAE from 0.05172 to 0.04931, force cosine from 0.26900 to 0.29511, and force-magnitude error from 0.08147 to 0.07597, while energy MAE increases from 0.54533 to 0.56208. Appendix \ref{app:controlled_replacements} gives the matched settings and full result.}

\paragraph{Expressiveness} Following \citep{joshi2023expressive}, we employ evaluation metrics that distinguish $n$-fold symmetric structures to precisely assess EST's expressiveness. We first construct a single-layer message-passing layer based on a EST-only structure, consistent with tensor product-based operations (TFN, MACE). Additionally, we remove the steerable FFN for a clear comparison. As demonstrated in \citep{dym2021on,joshi2023expressive}, the expressiveness of tensor product-based models is limited by the maximum degree $L$, failing to perfectly distinguish $n$-fold symmetric structures when $n > L$. From Table \ref{tab:rotsym}, we observe that EST significantly breaks through this theoretical boundary. It allows the model to distinguish symmetric structures even with very high fold (e.g., using 1-degree EST for 100-fold structures).
\begin{table}[htbp]
\caption{Experiments on Rotationally symmetric structures. } 
    \label{tab:rotsym}
        \centering
        \resizebox{\linewidth}{!}{
        \begin{tabular}{lccc|lcccc}
            \toprule
            GNN Layer & 3 fold & 10 fold & 100 fold & GNN Layer & 2 fold & 3 fold & 5 fold & 10 fold \\
            \midrule
            EST$_{L=1}$ & \cellcolor{green!10} \textbf{100.0 {\scriptsize ± 0.0}} & \cellcolor{green!10} \textbf{100.0 {\scriptsize ± 0.0}} & \cellcolor{green!10} \textbf{100.0 {\scriptsize ± 0.0}} & TFN/MACE$_{L=1}$ & \cellcolor{red!10} 50.0 {\scriptsize ± 0.0} & \cellcolor{red!10} 50.0 {\scriptsize ± 0.0} & \cellcolor{red!10} 50.0 {\scriptsize ± 0.0} & \cellcolor{red!10} 50.0 {\scriptsize ± 0.0} \\
            EST$_{L=2}$  & \cellcolor{green!10} \textbf{100.0 {\scriptsize ± 0.0}} & \cellcolor{green!10} \textbf{100.0 {\scriptsize ± 0.0}} & \cellcolor{green!10} \textbf{100.0 {\scriptsize ± 0.0}} & TFN/MACE$_{L=2}$ & \cellcolor{green!10} \textbf{100.0 {\scriptsize ± 0.0}} & \cellcolor{red!10} 50.0 {\scriptsize ± 0.0} & \cellcolor{red!10} 50.0 {\scriptsize ± 0.0} & \cellcolor{red!10} 50.0 {\scriptsize ± 0.0} \\
            EST$_{L=3}$  & \cellcolor{green!10} \textbf{100.0 {\scriptsize ± 0.0}} & \cellcolor{green!10} \textbf{100.0 {\scriptsize ± 0.0}} & \cellcolor{green!10} \textbf{100.0 {\scriptsize ± 0.0}} & TFN/MACE$_{L=3}$ & \cellcolor{green!10} \textbf{100.0 {\scriptsize ± 0.0}} & \cellcolor{green!10} \textbf{100.0 {\scriptsize ± 0.0}} & \cellcolor{red!10} 50.0 {\scriptsize ± 0.0} & \cellcolor{red!10} 50.0 {\scriptsize ± 0.0} \\
            EST$_{L=5}$  & \cellcolor{green!10} \textbf{100.0 {\scriptsize ± 0.0}} & \cellcolor{green!10} \textbf{100.0 {\scriptsize ± 0.0}} & \cellcolor{green!10} \textbf{100.0 {\scriptsize ± 0.0}} & TFN/MACE$_{L=5}$ & \cellcolor{green!10} \textbf{100.0 {\scriptsize ± 0.0}} & \cellcolor{green!10} \textbf{100.0 {\scriptsize ± 0.0}} & \cellcolor{green!10} \textbf{100.0 {\scriptsize ± 0.0}} & \cellcolor{red!10} 50.0 {\scriptsize ± 0.0} \\
            EST$_{L=10}$  & \cellcolor{green!10} \textbf{100.0 {\scriptsize ± 0.0}} & \cellcolor{green!10} \textbf{100.0 {\scriptsize ± 0.0}} & \cellcolor{green!10} \textbf{100.0 {\scriptsize ± 0.0}} & TFN/MACE$_{L=10}$ & \cellcolor{green!10} \textbf{100.0 {\scriptsize ± 0.0}} & \cellcolor{green!10} \textbf{100.0 {\scriptsize ± 0.0}} & \cellcolor{green!10} \textbf{100.0 {\scriptsize ± 0.0}} & \cellcolor{green!10} \textbf{100.0 {\scriptsize ± 0.0}} \\
            \bottomrule
        \end{tabular}
        }
    \vskip -0.3in
\end{table}
\begin{figure}[t]
    \centering
    \includegraphics[width=0.62\linewidth]{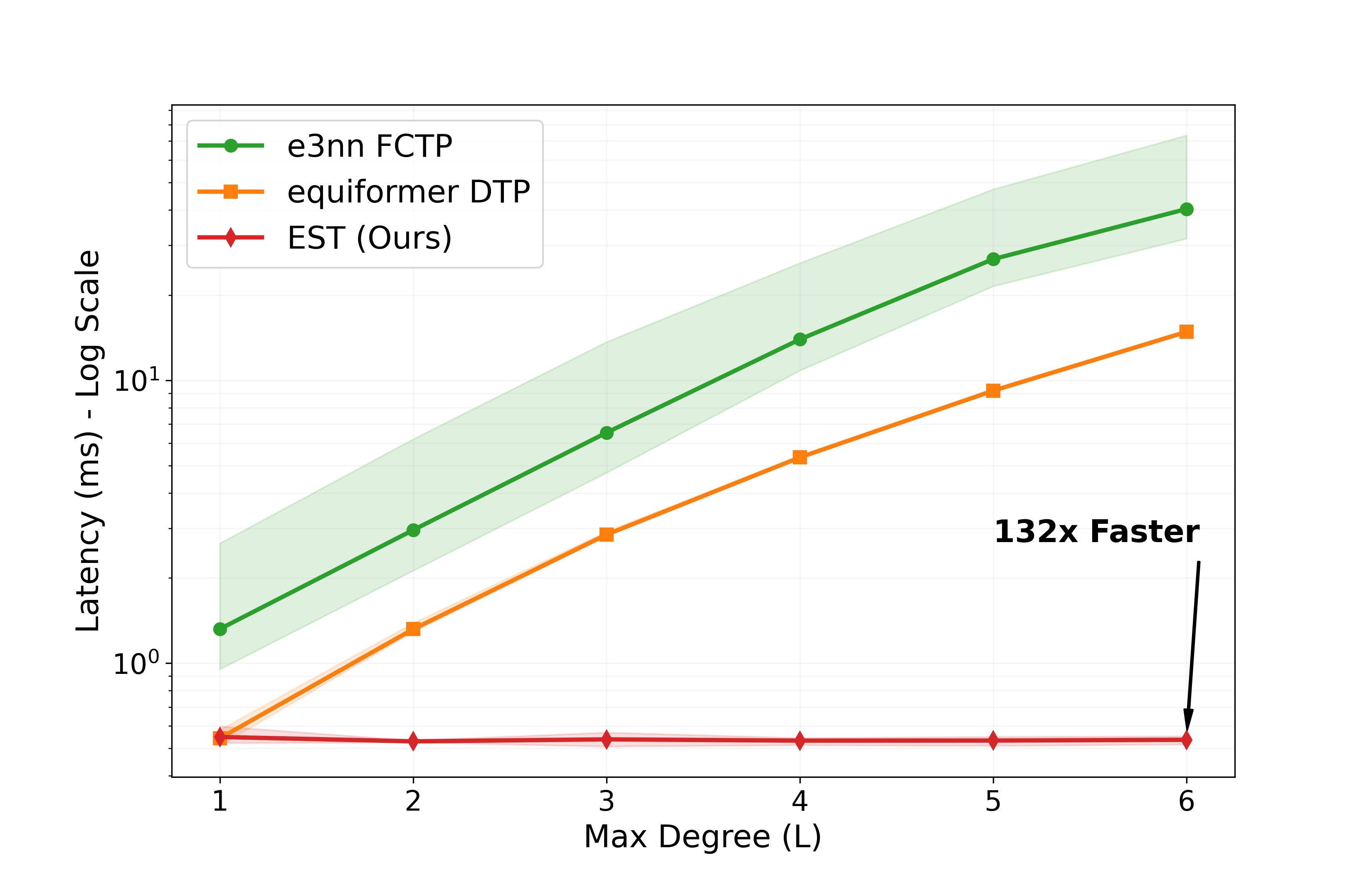}
    \caption{\textbf{Scaling analysis of inference latency.} EST, FCTP, and DTP are compared across maximum degree $L$ and channel dimension $C$. Curves show the mean over $C\in\{32,64,128,256,512\}$ and shaded regions show the range. EST achieves a 132$\times$ speedup over FCTP at $L=6,C=512$.}
    \label{fig:latency}
\end{figure}
\paragraph{Equivariance} \revision{Finite EST is approximately equivariant. We measure the mean absolute output difference before and after random rotations on 1,000 QM9 molecules using untrained 1--6 layer EST-only networks. Table \ref{tab:equivariance} shows that error decreases with more uniform or denser sampling; a trained 6-layer model gives $1.8\times10^{-5}\pm0.7\times10^{-5}$ averaged across QM9 properties. Independent geometric metrics in Appendix \ref{app:sampling_uniformity} confirm that FL and repulsion-refined FL are substantially more uniform than the e3nn grid.}
\begin{table}[tbp]
    \small
    \centering
    \caption{Equivariance evaluation with different sampling strategies.}
    \label{tab:equivariance}
    \begin{tabular}{l|cccccc}
        \toprule
         Sampling Strategy/Number & 1-layer & 2-layer & 3-layer & 4-layer & 5-layer & 6-layer \\
         \midrule
         FL / $S=64$, w/o optim. & 0.0010 & 0.0011 & 0.0010 & 0.0018 & 0.0022 & 0.0024 \\
         FL / $S=64$, w optim. & 0.0002 & 0.0001 & 0.0003 & 0.0003 & 0.0002 & 0.0002 \\
         FL / $S=256$, w/o optim. & 0.0002 & 0.0002 & 0.0001 & 0.0002 & 0.0001 & 0.0003 \\
         \hline
         e3nn / $S=210$, w/o optim. & 0.0084 & 0.1006 & 0.0366 & 0.0593 & 0.0046 & 0.0199 \\
         \bottomrule
    \end{tabular}
    \vskip -0.1in
\end{table}
\revision{\paragraph{Sampling-resolution guidance} Appendix \ref{app:resolution_tradeoff} jointly reports accuracy, equivariance error, runtime, and peak memory. At $S=128$, EST uses 2,538 MiB versus 1,624 MiB for the baseline; $S=512$ uses 12,424 MiB while improving $\alpha$ MAE by only 0.0003. We recommend $S=\max\{64,(2L)^2\}$, increasing it only when stricter equivariance or scarce-data regularization justifies the quadratic attention-memory cost.}

\paragraph{Latency} The theoretically analysis of EST's complexity is privided in Appendix \ref{app:complexity}. Here, we evaluate the computational efficiency of our EST against the Fully Connected Tensor Product (FCTP) and the Depthwise Tensor Product (DTP). We benchmarked the feed-forward latency of these operators across varying channel dimensions ($C$) and maximum degree ($L$). All experiments were conducted on a server equipped with an Intel(R) Xeon(R) Platinum 8558 CPU and an NVIDIA H200 GPU. Detailed results are recorded in Table \ref{tab:latency}. As illustrated in Fig. \ref{fig:latency}, the computational overhead of FCTP and DTP grows exponentially with $L$, whereas the EST remains remarkably stable. Furthermore, FCTP exhibits high sensitivity to channel scaling. Its latency increases sharply as $C$ expands from 32 to 512. In contrast, both DTP and EST demonstrate superior scalability with respect to channel depth. Notably, at $C=512$ and $L=6$, the EST achieves an approximate 132$\times$ speedup over the traditional FCTP. This efficiency stems not only from our architectural design but also from the integration of deep learning engineering and infrastructure, e.g., hardware-friendly Transformations.

%% file: chapter/conclusion.tex
\section{Conclusion}\label{sec:conclusion}
\revision{We present EST, a plug-and-play spherical Transformer that adds cross-orientation interactions to equivariant molecular models. The continuous operator is exactly equivariant on a rotation-closed domain, whereas the finite implementation is approximately equivariant. A controlled replacement in SCN isolates gains from cross-orientation attention, and experiments on OC20, MPtrj, and QM9 show the largest benefits on complex tasks with modeling headroom. NVE, relaxation, and vibrational tests further support applying EST in dynamical and structural workflows, particularly with conservative gradient forces. Limitations are the $O(S^2)$ attention memory, finite-sampling equivariance error, increased non-conservativity with a direct force head, and attenuated gains on deeper or saturated backbones. The recommended $S=\max\{64,(2L)^2\}$ balances accuracy and equivariance against memory.}



%% file: chapter/appendix.tex
\newpage
\appendix

\section*{APPENDIX}
\begin{adjustwidth}{2cm}{}
\addcontentsline{toc}{section}{Appendix}
\startcontents[Appendix]
\printcontents[Appendix]{l}{1}{\setcounter{tocdepth}{2}}
\end{adjustwidth}






\input{chapter/appendix/mathematic}
\input{chapter/appendix/theorem_proof}
\input{chapter/appendix/rebuttal_evidence}
\input{chapter/appendix/supplementary_experiments}

\input{chapter/appendix/moe_details}

%% file: chapter/appendix/mathematic.tex
\section{The Mathematics}\label{app:spherical_harmonic}

\subsection{Message Passing Neural Networks} \label{app:mpnn}
Consider a graph $\mathcal{G} =(\mathcal{V},\mathcal{E})$, with nodes $v_i \in \mathcal{V}$ and edges $e_{ij} \in \mathcal{E}$. Each node $v_i$ has an embedding $\mathbf{x}_i$ and an attribute $\mathbf{z}_i$, and each edge $e_{ij}$ has an attribute $\mathbf{a}_{ij}$. The Message Passing Neural Network (MPNN) \citep{schutt2018schnet}, a specific type of GNN, updates node embeddings through a message block $\mathbf{M}(\cdot)$ and an update block $\mathbf{U}(\cdot)$ via the following steps:
\begin{equation}
\label{equ:messageblock}
    \mathbf{m}_{ij} = \mathbf{M}(\mathbf{x}_{i}, \mathbf{x}_{j}, \mathbf{z}_{i}, \mathbf{z}_{j}, \mathbf{a}_{ij}, \vec{\mathbf{r}}_{ij}), \quad \text{and} \quad \tilde{\mathbf{x}}_{i} = \mathbf{U}(\mathbf{x}_{i}, \sum_{j \in \mathcal{N}(i)} \mathbf{m}_{ij}),
\end{equation}
where we use $\vec{\mathbf{r}}_{ij}$ to denote the relative position of node $i$ and node $j$ in 3D space, and the neighborhood $\mathcal{N}(i)$ is typically defined by a cutoff radius: $\mathcal{N}(i) = \{j \mid \|\vec{\mathbf{r}}_{ij}\|\leq r_{cut}\}$. The node attribute $\mathbf{z}_{i}$ contains atomic information such as the atomic type, and the edge attribute $\mathbf{a}_{ij}$ contains atomic pair information such as distance and bond type. By stacking multiple message and update blocks, the final node embeddings can be used to model atomic interactions or represent molecular properties.

\subsection{The Mathematics of Spherical Harmonics}\label{app:spherical_harmonic_detail}
\subsubsection{The Properties of Spherical Harmonics}
The spherical harmonics $Y^{(l,m)}(\theta,\varphi)$ are the angular portion of the solution to Laplace's equation in spherical coordinates where azimuthal symmetry is not present.
Some care must be taken in identifying the notational convention being used.
In this entry, $\theta$ is taken as the polar (colatitudinal) coordinate with $\theta$ in $[0,\pi]$, and $\varphi$ as the azimuthal (longitudinal) coordinate with $\varphi$ in $[0,2\pi)$. 

Spherical harmonics satisfy the spherical harmonic differential equation, which is given by the angular part of Laplace's equation in spherical coordinates. If we define the solution of Laplace's equation as $F=\Phi(\varphi)\Theta(\theta)$, the equation can be transformed as:
\begin{equation}
    \frac{\Phi(\varphi)}{\sin \theta} \frac{d}{d \theta}\left(\sin \theta \frac{d \Theta}{d \theta}\right)+\frac{\Theta(\theta)}{\sin ^{2} \theta} \frac{d^{2} \Phi(\varphi)}{d \varphi^{2}}+l(l+1) \Theta(\theta) \Phi(\varphi)=0 
\end{equation}
Here we omit the derivation process and just show the result. The (complex-value) spherical harmonics are defined by:
\begin{equation}
 Y^{(l,m)}(\theta, \varphi) \equiv \sqrt{\frac{2 l+1}{4 \pi} \frac{(l-m) !}{(l+m) !}} P^{(l,m)}(\cos \theta) e^{i m \varphi},
\end{equation}
where $P^{(l,m)}(\cos \theta)$ is an associated Legendre polynomial.
Spherical harmonics are integral basis, which satisfy:
\begin{equation}\label{equ:cg2sh1}
\begin{array}{l}
\int_{0}^{2 \pi} \int_{0}^{\pi} Y^{(l_{1},m_{1})}(\theta, \varphi) Y^{(l_{2},m_{2})}(\theta, \varphi) Y^{(l_{3},m_{3})}(\theta, \varphi) \sin \theta d \theta d \varphi \\
=\sqrt{\frac{\left(2 l_{1}+1\right)\left(2 l_{2}+1\right)\left(2 l_{3}+1\right)}{4 \pi}}\left(\begin{array}{ccc}
l_{1} & l_{2} & l_{3} \\
0 & 0 & 0
\end{array}\right)\left(\begin{array}{ccc}
l_{1} & l_{2} & l_{3} \\
m_{1} & m_{2} & m_{3}
\end{array}\right),
\end{array}
\end{equation}
where $\left(\begin{array}{ccc}
l_{1} & l_{2} & l_{3} \\
m_{1} & m_{2} & m_{3}
\end{array}\right)$ is a Wigner 3j-symbol (which is related to the Clebsch-Gordan coefficients).
We list a few spherical harmonics which are:

\begin{equation}
\label{equ:app_sh}
\begin{aligned}
Y^{(0,0)}(\theta, \varphi) & =\frac{1}{2} \sqrt{\frac{1}{\pi}}, \\
Y^{(1,-1)}(\theta, \varphi) & =\frac{1}{2} \sqrt{\frac{3}{2 \pi}} \sin \theta e^{-i \varphi}, \\
Y^{(1,0)}(\theta, \varphi) & =\frac{1}{2} \sqrt{\frac{3}{\pi}} \cos \theta, \\
Y^{(1,1)}(\theta, \varphi) & =\frac{-1}{2} \sqrt{\frac{3}{2 \pi}} \sin \theta e^{i \varphi}, \\
Y^{(2,-2)}(\theta, \varphi) & =\frac{1}{4} \sqrt{\frac{15}{2 \pi}} \sin { }^{2} \theta e^{-2 i \varphi}, \\
Y^{(2,-1)}(\theta, \varphi) & =\frac{1}{2} \sqrt{\frac{15}{2 \pi}} \sin \theta \cos \theta e^{-i \varphi}, \\
Y^{(2,0)}(\theta, \varphi) & =\frac{1}{4} \sqrt{\frac{5}{\pi}}\left(3 \cos ^{2} \theta-1\right), \\
Y^{(2,1)}(\theta, \varphi) & =\frac{-1}{2} \sqrt{\frac{15}{2 \pi}} \sin \theta \cos \theta e^{i \varphi}, \\
Y^{(2,2)}(\theta, \varphi) & =\frac{1}{4} \sqrt{\frac{15}{2 \pi}} \sin { }^{2} \theta e^{2 i \varphi}, \\
\end{aligned}
\end{equation}
In this work, we use the real-value spherical harmonics rather than the complex-value one, which can be written as : 
\begin{equation}
\label{equ:app_sh_real}
\begin{aligned}
   Y^{0,0}(\theta,\varphi) &= \sqrt{\frac{1}{4\pi}},\\
   Y^{(1,-1)}(\theta,\varphi) &= \sqrt{\frac{3}{4\pi}}\sin\varphi\sin\theta,\\
   Y^{(1,0)}(\theta,\varphi) &= \sqrt{\frac{3}{4\pi}}\cos\theta,\\
   Y^{(1,1)}(\theta,\varphi) &= \sqrt{\frac{3}{4\pi}}\cos\varphi\sin\theta,\\
   Y^{(2,-2)}(\theta,\varphi) &= \sqrt{\frac{15}{16\pi}}\sin(2\varphi)\sin^2\theta ,\\
   Y^{(2,-1)}(\theta,\varphi) &= \sqrt{\frac{15}{4\pi}}\sin\varphi\sin\theta\cos\theta,\\
   Y^{(2,0)}(\theta,\varphi) &= \sqrt{\frac{5}{16\pi}}(3\cos^2\theta - 1),\\
   Y^{(2,1)}(\theta,\varphi) &= \sqrt{\frac{15}{4\pi}}\cos\varphi\sin\theta\cos\theta,\\
   Y^{(2,2)}(\theta,\varphi) &= \sqrt{\frac{15}{16\pi}}\cos(2\varphi)\sin^2\theta. \\
\end{aligned}
\end{equation}

\subsubsection{Fourier transformation over $\mathbb{S}^2$} 
In the main paper, we show that any square-integrable function $f(\cdot)$ can thus be expanded as a linear combination of spherical harmonics:
\begin{equation}\label{app_equ:inverse_fourier}
    f(\vec{\mathbf{p}}) = \sum_{l=0}^{\infty} \sum^{l}_{m=-l} \mathbf{x}^{(l,m)}Y^{(l,m)}(\vec{\mathbf{p}}),
\end{equation}
where $\vec{\mathbf{p}} =(\theta,\varphi)$ denotes the orientations, like what we do in the main paper. The coefficient $\mathbf{x}^{(l,m)}$ can be obtained by the inverse transformation over $\mathbb{S}^2$, which is
\begin{equation}\label{app_equ:fourier}
    \mathbf{x}^{l,m} = \int_{\Omega} f(\vec{\mathbf{p}}) Y^{(l,m)*}(\vec{\mathbf{p}}) d\Omega = \int_{0}^{2\pi} d\varphi \int_{0}^{\pi} d\theta \sin \theta f(\vec{\mathbf{p}}) Y^{(l,m)*}(\vec{\mathbf{p}}).  
\end{equation}
Using the fact $ \mathbf{Y}^{l}(\mathbf{R}\vec{\mathbf{p}}) = \mathbf{D}^{l}(\mathbf{R}) \mathbf{Y}^{l}(\vec{\mathbf{p}}), 
$ and equation \eqref{app_equ:inverse_fourier}, we know 
\begin{equation}
f(\mathbf{R} \vec{\mathbf{p}}) = \sum_{l=0}^{\infty} \sum^{l}_{m=-l} \mathbf{x}^{(l,m)}Y^{(l,m)}(\mathbf{R}\vec{\mathbf{p}}) = \sum_{l=0}^{\infty} \sum^{l}_{m=-l}  \mathbf{x}^{(l,m)} \mathbf{D}Y^{(l,m)}(\vec{\mathbf{p}}).
\end{equation}
Therefore, we can get the conclusion that spatial representation $f(\mathbf{R}\vec{\mathbf{P}})$ and $f(\vec{\mathbf{P}})$ is steerable, which can be represented by
\begin{equation}
\label{equ:equivariance_fft}
f(\mathbf{R}\vec{\mathbf{p}}) = \mathbf{D}^{-1} \mathbf{x} = \mathbf{D}^{T} \mathbf{x}. 
\end{equation}

\subsubsection{The Relationship Between Spherical Harmonics and Wigner-D Matrix}
One can also think of standard spherical harmonics as functions $\tilde{Y}^{(l)}_m:\text{SO}(3) \rightarrow \mathbb{R}$ on SO$(3)$ that are invariant to a sub-group of rotations via
\begin{equation*}
Y^{(l,m)}(\mathbf{n}_{\alpha,\beta}) = Y^{(l,m)}(\mathbf{R}_{\alpha,\beta,\gamma}\mathbf{n}_x)=:\tilde{Y}^{(l,m)}(\mathbf{R}_{\alpha,\beta,\gamma}) \ .
\end{equation*}
Then, by definition, $\tilde{Y}{(l,m)}$ is invariant with respect to rotation angle $\gamma$, i.e.., $\forall_{\gamma \in [0,2 \pi)}: \;\; \tilde{Y}^{(l,m)}(\mathbf{R}_{\alpha,\beta,\gamma}) = \tilde{Y}^{(l,m)}(\mathbf{R}_{\alpha,\beta,0})$.  
This viewpoint of regarding the spherical harmonics as $\gamma$-invariant functions on O($3$) helps us to draw the connection to the Wigner-D functions $D_{mn}^{(l)}$ that make up the $2l+1\times2l+1$ elements of the Wigner-D matrices. Namely, the $n=0$ column of Wigner-D functions are also $\gamma$-invariant and, in fact, correspond (up to a normalisation factor) to the spherical harmonics via
\begin{equation}
Y^{(l,m)}(\mathbf{n}_{\alpha,\beta}) = \frac{1}{\sqrt{2l+1}} D^{(l)}_{m0}(\mathbf{R}_{\alpha,\beta,\gamma}) \ .
\end{equation} 

\subsubsection{Equivariance of Clebsch-Gordan Tensor Product}
The Clebsch-Gordan Tensor Product shows a strict equivariance for different
group representations, which make the mixture representations transformed equivariant based on Wigner-D matrices. 
We use $\mathbf{D}^{l,m'm}$ to denote the element of Wigner-D matrix in the degree $l$.
The Clebsch-Gordan coefficient satisfies:
\begin{equation}
\begin{array}{c}
\sum_{m_{1}^{\prime}, m_{2}^{\prime}} \mathcal{C}_{\left(l_{1}, m_{1}^{\prime}\right)\left(l_{2}, m_{2}^{\prime}\right)}^{\left(l_{0}, m_{0}\right)} \mathbf{D}^{(l_{1},m_{1}^{\prime} m_{1})}(g) \mathbf{D}^{(l_{2},m_{2}^{\prime} m_{2})}(g) \\
=\sum_{m_{0}^{\prime}} \mathbf{D}^{(l_{0},m_{0}^{\prime} m_{0})}(g)  \mathcal{C}_{\left(l_{1}, m_{1}\right)\left(l_{2}, m_{2}\right)}^{\left(l_{0}, m_{0}^{\prime}\right)} \\
\end{array}
\end{equation}
Therefore, the spherical harmonics can be combined equivariantly by CG Tensor Product:
\begin{equation}
\label{equ:wignercg}
    \begin{aligned}
&CG\left( \sum_{m_{1}^{\prime}} \mathbf{D}^{(l_{1},m_{1} m_{1}^{\prime})}(g) Y^{(l_{1},m_{1}^{\prime})}, \sum_{m_{2}^{\prime}}  \mathbf{D}^{(l_{2},m_{2} m_{2}^{\prime})}(g) Y^{(l_{2},m_{2}^{\prime})} \right) \\
=&\sum_{m_{1}, m_{2}} \mathcal{C}_{\left(l_{1}, m_{1}\right)\left(l_{2}, m_{2}\right)}^{\left(l_{0}, m_{0}\right)} \sum_{m_{1}^{\prime}} \mathbf{D}^{(l_{1},m_{1} m_{1}^{\prime})}(g) Y^{(l_{1},m_{1}^{\prime})} \sum_{m_{2}^{\prime}}  \mathbf{D}^{(l_{2},m_{2} m_{2}^{\prime})}(g) Y^{(l_{2},m_{2}^{\prime})} \\
=&\sum_{m_{0}^{\prime}}  \mathbf{D}^{(l_{0},m_{0} m_{0}^{\prime})}(g) \sum_{m_{1}, m_{2}} \mathcal{C}_{\left(l_{1}, m_{1}\right)\left(l_{2}, m_{2}\right)}^{\left(l_{0} m_{0}^{\prime}\right)} Y^{(l_{1},m_{1}^{\prime})} Y^{(l_{2},m_{2}^{\prime})}
\\
=&\sum_{m_{0}^{\prime}} \mathbf{D}^{(l_{0},m_{0} m_{0}^{\prime})}(g) CG\left(Y^{(l_{1},m_{1}^{\prime})}, Y^{(l_{2},m_{2}^{\prime})} \right) .
\end{aligned}
\end{equation}
Here, we omit the input argument of the spherical harmonics, which can represent any direction on the sphere. Equation \eqref{equ:wignercg} represents a relationship between scalar. If we transform the scalar to vector or matrix like what we do in equation \eqref{equ:Wigner-D}, equation \eqref{equ:wignercg} is equal to
\begin{equation}
\label{equ:CG_Wigner}
    (\mathbf{D}^{l_{1}} \mathbf{u} \otimes \mathbf{D}^{l_{2}} \mathbf{v})^{l_{0}} = \mathbf{D}^{l_{0}}(\mathbf{u} \otimes \mathbf{v})^{l_{0}}.
\end{equation}
The tensor CG product mixes two representations to a new representation under special rule $|l_{1}-l_{2}| \le l \le (l_{1} + l_{2})$.
For example, 1.two type-$0$ vectors will only generate a type-$0$ representations; 2.type-$l_{1}$ and type-$l_{2}$ can generate type-$l_{1}+l_{2}$ vector at most.
Note that some widely-used products are related to tensor product: scalar product ($l_1=0$, $l_2=1$, $l=1$), dot product ($l_1=1$, $l_2=1$, $l=0$) and cross product ($l_1=1$, $l_2=1$, $l=1$). 
However, for each element with $l>0$, there are multi mathematical operation for the connection with weights.
The relation between number of operations and degree is quadratic. 
Thus, as degree increases, the amount of computation increases significantly, making calculation of the CG tensor product slow for higher order irreps. 
This statement can be proven by the implementation of e3nn (o3.FullyConnectedTensorProduct).

\subsubsection{Learnable Parameters in Tensor Product}
Previous works utilize the e3nn library~\citep{e3nn} to implement the corresponding tensor product. It is crucial to emphasize that the formulation of CG tensor product is devoid of any learnable parameters, as CG coefficients remain constant. In the context of e3nn, learnable parameters are introduced into each path, represented as $w(\mathbf{u}^{l_{1}} \otimes \mathbf{v}^{l_{2}})$. Importantly, these learnable parameters will not destory the equivariance of each path. However, they are limited in capturing directional information. In equivariant models, the original CG tensor product primarily captures directional information. We have previously mentioned our replacement of the CG tensor product with learnable modules. It is worth noting that our focus lies on the CG coefficients rather than the learnable parameters in the e3nn implementation. 

\subsubsection{Gate Activation and Normalization}
\textbf{Gate Activation.} In equivariant models, the Gate activation combines two sets of group representations. The first set consists of scalar steerable vector ($l=0$), which are passed through standard activation functions such as sigmoid, ReLU and SiLU. The second set comprises higher-order steerable vector (($l>0$)), which are multiplied by an additional set of scalar steerable vector that are introduced solely for the purpose of the activation layer. These scalar steerable vector are also passed through activation functions.

\textbf{Normalization.} Normalization is a technique commonly used in neural networks to normalize the activations within each layer. It helps stabilize and accelerate the training process by reducing the internal covariate shift, which refers to the change in the distribution of layer inputs during training. 

The normalization process involves computing the mean and variance across the channels. In equivariant normalization, the variance is computed using the root mean square value of the L2-norm of each type-$l$ vector. Additionally, this normalization removes the mean term. The normalized activations are then passed through a learnable affine transformation without a learnable bias, which enables the network to adjust the mean and variance based on the specific task requirements.

\subsection{Relationship Between Expressive Power and Equivariant Operations}
In~\citep{dym2021on}, Theorem 2 establishes the universality of equivariant networks based on the TFN structure:

\textbf{Theorem.} For all $ n \in \mathbb{N}$, $\mathbf{l}_T \in \mathbb{N_{+}}^{*} $,

\begin{itemize}
    \item 1. For $D \in \mathbb{N_{+}}$, every G-equivariant polynomial $p$ : $\mathbb{R}^{3 \times n}$ $ \to W_{\mathbf{l}_T}^{n}$ of degree $D$ is in $F^{TFN}_{C(D), D}$. \\

    \item 2. Every continuous G-equivariant function can be approximated uniformly on compact sets by functions in $ \cup_{D \in \mathbb{N_{+}}}F^{TFN}_{C(D),D} $.  \\
\end{itemize}
Here, $n$ represents the number of input points (or nodes), $\mathbf{l}_T$ represents the degree of the approximated G-equivariant function, $C$ represents the number of channels, and $D$ represents the degree of the TFN (Tensor Field Network) structure, which is equivalent to the term $l$ used in our method. The TFN structure consists of two layers, including convolution and self-interaction. Self-interaction involves equivariant linear functions. The convolution operation calculates the CG tensor product between different steerable representations, which is a fundamental operation for transforming directional information. Most equivariant models based on group representations use a similar approach (CG tensor product) to capture directional features. Therefore, the theorem mentioned above also applies to building blocks based on CG tensor products, such as SEGNN~\citep{brandstetter2021geometric} and Equiformer~\citep{liao2023equiformer}.

It is important to note that achieving an infinite degree in practice is not feasible. However, equivariant models based on group representations can enhance their expressive power by increasing the number of maximal degrees~\citep{dym2021on}. In their evaluation of expressive power, as presented in~\citep{joshi2023expressive}, the authors utilize the GWL (geometric Weisfeiler-Leman) graph isomorphism test. In Table 2 of their work, it is evident that equivariant models with a maximal degree denoted as $L$ are incapable of distinguishing $n$-fold symmetric structures when $n$ exceeds the value of $L$.

%% file: chapter/appendix/theorem_proof.tex
\section{Proofs and Details For Section \ref{sec:model}}\label{app:proof}
\subsection{Proof for Equivariance of EST}\label{app:proof_equivariance}
\revision{We prove Theorem \ref{theorem:equivariance} for a rotation-closed continuous domain $\Omega$ with rotation-invariant measure.}
\begin{proof}
Recall that FT can be represented by 
\begin{equation}
    f(\vec{\mathbf{p}}) = \sum_{l=0}^{L}\sum_{m=-l}^{l}x^{(l,m)}Y^{(l,m)}(\vec{\mathbf{p}}) = (\mathbf{x}^{(0\rightarrow L)})^{T}\mathbf{Y}^{(0\rightarrow L)}(\vec{\mathbf{p}}).
\end{equation}
When we apply a random rotation $\mathbf{D}^{-1}$ to $\mathbf{x}$, we obtain
\begin{align}
    & (\mathbf{D}^{-1} \mathbf{x}^{(0\rightarrow L)})^{T}\mathbf{Y}^{(0\rightarrow L)}(\vec{\mathbf{p}}) \\
    = & (\mathbf{x}^{(0\rightarrow L)})^{T} \mathbf{D}\mathbf{Y}^{(0\rightarrow L)}(\vec{\mathbf{p}}) \\
    = & (\mathbf{x}^{(0\rightarrow L)})^{T} \mathbf{Y}^{(0\rightarrow L)}(\mathbf{R}\vec{\mathbf{p}}) \\
    = & f(\mathbf{R} \vec{\mathbf{p}}),
\end{align}
where $\mathbf{R}$ and $\mathbf{D}$ share the same transformation parameters.

Given a spatial representation $f(\vec{\mathbf{p}})$ transformed from a steerable representation $\mathbf{x}$, spherical attention can be represented as:
\begin{equation}
    \tilde{f}(\vec{\mathbf{p}}_{1}) = \int_{\vec{\mathbf{p}}_{2} \in \Omega}a\big(f^{Q}(\vec{\mathbf{p}}_{1}), f^{K}(\vec{\mathbf{p}}_{2})\big)f^{V}(\vec{\mathbf{p}}_{2})d\vec{\mathbf{p}}_{2},
\end{equation}
where $a(\cdot)$ denotes the operation to compute attention coefficients. Terms $Q,K,V$ denote three independent linear transformations. When the origin representation $\mathbf{x}$ is rotation by a Wigner-D matrix $\mathbf{D}^{-1}$, the representation $f(\vec{\mathbf{p}})$ is transformed to $f(\mathbf{R} \vec{\mathbf{p}})$. Therefore, the attention results are changed to
\begin{equation}
\begin{aligned}
\label{equ:proof_equivariance}
    &\int_{\mathbf{R}\vec{\mathbf{p}}_{2} \in \Omega}a\big(f^{Q}(\mathbf{R}\vec{\mathbf{p}}_{1}), f^{K}(\mathbf{R}\vec{\mathbf{p}}_{2})\big)f^{V}(\mathbf{R}\vec{\mathbf{p}}_{2})d\mathbf{R}\vec{\mathbf{p}}_{2} \\
    =& \int_{\mathbf{R}\vec{\mathbf{p}}_{2} \in \Omega} \frac{\exp\big(f^{Q}(\mathbf{R}\vec{\mathbf{p}}_{1})*f^{K}(\mathbf{R}\vec{\mathbf{p}}_{2})\big)}{\int_{\mathbf{R}\vec{\mathbf{p}}_{3}\in\Omega}\exp\big(f^{Q}(\mathbf{R}\vec{\mathbf{p}}_{1})*f^{K}(\mathbf{R}\vec{\mathbf{p}}_{3})\big)d\mathbf{R}\vec{\mathbf{p}}_{3}}
    f^{V}(\mathbf{R}\vec{\mathbf{p}}_{2})d\mathbf{R}\vec{\mathbf{p}}_{2} \\
    =& \int_{\mathbf{R}\vec{\mathbf{p}}_{2} \in \Omega} \frac{\exp\big(f^{Q}(\mathbf{R}\vec{\mathbf{p}}_{1})*f^{K}(\mathbf{R}\vec{\mathbf{p}}_{2})\big)}{\int_{\vec{\mathbf{p}}_{3}\in\Omega}\exp\big(f^{Q}(\mathbf{R}\vec{\mathbf{p}}_{1})*f^{K}(\vec{\mathbf{p}}_{3})\big)d\vec{\mathbf{p}}_{3}}
    f^{V}(\mathbf{R}\vec{\mathbf{p}}_{2})d\mathbf{R}\vec{\mathbf{p}}_{2} \\
    =& \int_{\vec{\mathbf{p}}_{2} \in \Omega} \frac{\exp\big(f^{Q}(\mathbf{R}\vec{\mathbf{p}}_{1})*f^{K}(\vec{\mathbf{p}}_{2})\big)}{\int_{\vec{\mathbf{p}}_{3}\in\Omega}\exp\big(f^{Q}(\mathbf{R}\vec{\mathbf{p}}_{1})*f^{K}(\vec{\mathbf{p}}_{3})\big)d\vec{\mathbf{p}}_{3}}
    f^{V}(\vec{\mathbf{p}}_{2})d\vec{\mathbf{p}}_{2} \\
    =& \int_{\vec{\mathbf{p}}_{2} \in \Omega}a\big(f^{Q}(\mathbf{R}\vec{\mathbf{p}}_{1}), f^{K}(\vec{\mathbf{p}}_{2})\big)f^{V}(\vec{\mathbf{p}}_{2})d\vec{\mathbf{p}}_{2} \\
    =& \tilde{f}(\mathbf{R}\vec{\mathbf{p}}_{1}) .\\
\end{aligned}
\end{equation}
\revision{The changes of variables above are valid because $\mathbf R\Omega=\Omega$ and rotations preserve surface measure. Therefore, the output of spherical attention remains steerable after rotation. If we transform $\tilde{f}(\mathbf{R}\vec{\mathbf{p}}_{1})$ into its steerable representation, denoted as $\tilde{\mathbf{x}}$, the following relationship holds:}
\begin{equation}
    \tilde{\mathbf{x}} = \mathbf{D}^{-1} \mathcal{F}^{-1}\big(\mathrm{SA}(\mathcal{F}(\mathbf{x}))\big) = \mathcal{F}^{-1}\big(\mathrm{SA}(\mathcal{F}(\mathbf{D}^{-1}\mathbf{x}))\big)
\end{equation}
\revision{Thus spherical attention is equivariant on the closed continuous domain. A shared pointwise spherical FFN also commutes with rotation because it applies the same function at every orientation, completing the proof for the continuous EST operator.}


In the derivation above, the rotational term is effectively omitted in the integration limits because the sphere is a symmetric set that is invariant under any rotation. Specifically, the continuous spherical surface $\Omega$ satisfies $\mathbf{R} \Omega \equiv \Omega$ for any $\mathbf{R} \in SO(3)$. This implies that for any direction $\vec{\mathbf{p}} \in \Omega$, the rotated point $\mathbf{R}\vec{\mathbf{p}}$ remains in $\Omega$, and this mapping is bijective. Since the Jacobian determinant of a rotation matrix is 1, the integral measure is invariant, i.e., $d(\mathbf{R}\vec{\mathbf{p}}) = d\vec{\mathbf{p}}$. Thus, integrating over the rotated coordinates yields the same result as integrating over the original coordinates. 

To illustrate this, consider a 2D analog: a periodic function on a circle $y = f(\theta)$. The integral over the circle is $\int^{2\pi}_{0}f(\theta)\,d\theta$. If we rotate the domain by a constant $C$, the function becomes $f(\theta + C)$. Due to the periodicity of the circle ($f(2\pi + \theta) = f(\theta)$), the integral $\int^{2\pi}_{0}f(\theta + C)\,d\theta$ yields the same result as the original integral. The 3D spherical case follows the same principle: rotation changes the relative positions of points, but the integral over the entire closed surface remains invariant.

\paragraph{Conclusion of Proof:}
Since $\tilde{f}(\mathbf{R}\vec{\mathbf{p}}_{1})$ is the rotated output, if we transform it back into its steerable representation $\tilde{\mathbf{x}}$, the following relationship holds:
\begin{equation}
    \tilde{\mathbf{x}} = \mathbf{D}^{-1} \mathcal{F}^{-1}\big(\mathrm{SA}(\mathcal{F}(\mathbf{x}))\big) = \mathcal{F}^{-1}\big(\mathrm{SA}(\mathcal{F}(\mathbf{D}^{-1}\mathbf{x}))\big).
\end{equation}
Thus, the equivariance of spherical attention holds. Moreover, the spherical FFN is inherently equivariant because the function $\tilde{f}(\vec{\mathbf{p}}_{1}) = \mathrm{FFN}\big(f(\vec{\mathbf{p}}_{1})\big)$ operates locally on each orientation. Therefore, the entire EST framework is equivariant.
\end{proof}

\paragraph{Discussion on Finite Sampling and Equivariance:}
\revision{Theorem \ref{theorem:equivariance} establishes exact equivariance only for the closed continuous domain. Practical EST replaces the integral by a finite quadrature and is therefore approximately equivariant for arbitrary rotations.}

In the discrete case, the attention operation in equation \eqref{equ:proof_equivariance} is approximated as:
\begin{equation}
\label{equ:discrete_sa}
    \tilde{f}(\vec{\mathbf{p}}_{1}) = \sum_{\vec{\mathbf{p}}_{2}} a\big(f^{Q}(\vec{\mathbf{p}}_{1}), f^{K}(\vec{\mathbf{p}}_{2})\big)f^{V}(\vec{\mathbf{p}}_{2}). 
\end{equation}
After introducing random rotations, this becomes:
\begin{equation}
\label{equ:discrete_sa_rota}
    \tilde{f}(\mathbf{R} \vec{\mathbf{p}}_{1}) = \sum_{\vec{\mathbf{p}}_{3}} a\big(f^{Q}(\mathbf{R} \vec{\mathbf{p}}_{1}), f^{K}(\vec{\mathbf{p}}_{3})\big)f^{V}(\vec{\mathbf{p}}_{3}). 
\end{equation}
\revision{A finite point set cannot be closed under all of $SO(3)$. Exact discrete equivariance holds only for rotations in the grid-symmetry subgroup that permute all samples; other rotations incur quadrature and interpolation mismatch.}

\revision{For a grid-symmetry rotation, equation \eqref{equ:discrete_sa_rota} equals equation \eqref{equ:discrete_sa} up to permutation. Uniformity alone does not create exact closure; it reduces the numerical mismatch for rotations outside this subgroup.}

To mitigate this, we employ approximately uniform sampling methods, such as Fibonacci Lattices. Furthermore, as established in prior literature, neural networks are capable of approximating continuous equivariant functions. Rotations not strictly covered by the discrete sampling grid are handled via training, allowing the model to learn approximate equivariance. Our experiments confirm that the equivariance error reduces significantly after training without requiring explicit data augmentation.

\begin{figure}[htbp]
\centering
    \begin{subfigure}[b]{0.48\linewidth}
        \includegraphics[width=\linewidth, trim=0cm 0cm 0cm 1cm,clip]{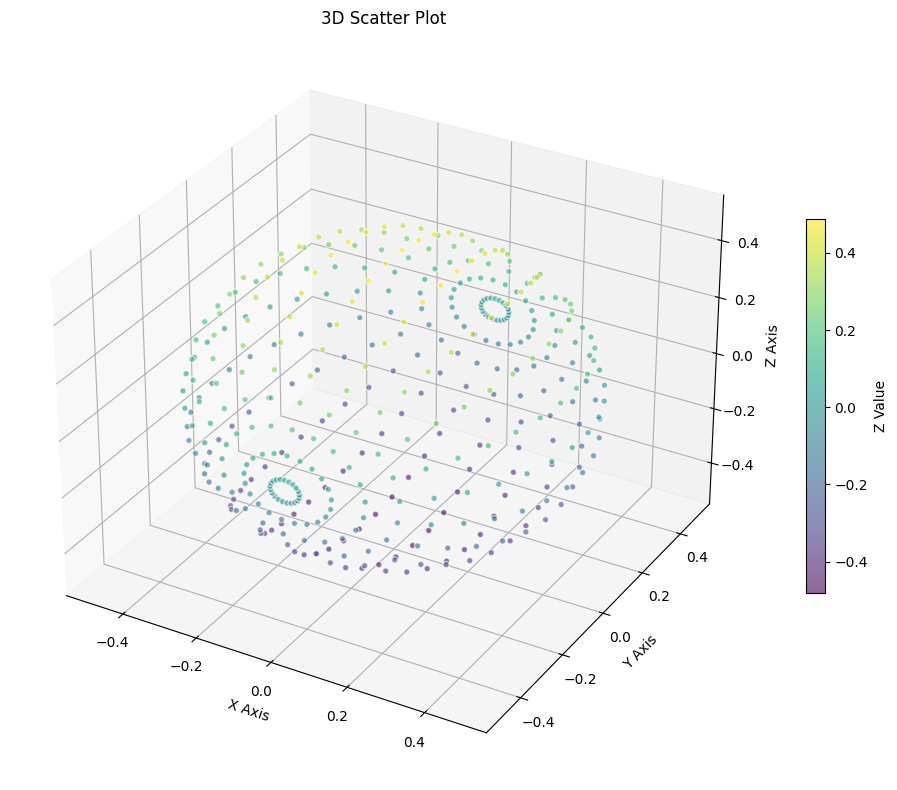}
        \caption{e3nn Sampling}
    \end{subfigure}
    \hfill
    \begin{subfigure}[b]{0.48\linewidth}
        \includegraphics[width=\linewidth, trim=0cm 0cm 0cm 1cm,clip]{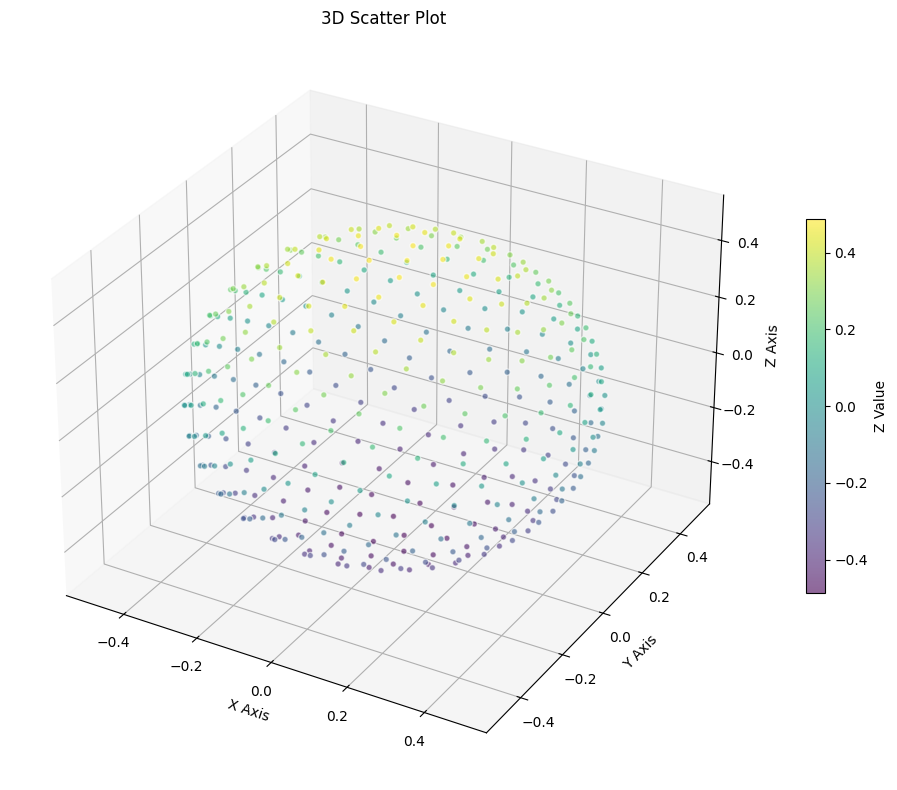}
        \caption{Fibonacci Lattices Sampling}
    \end{subfigure}
    \caption{Two spherical sampling strategies.}
    \label{fig:sampling}
\end{figure}

\paragraph{Sampling strategies} Most previous works use the e3nn implementation for spherical Fourier transform. However, it significantly destroy the uniformity of spherical sampling, which is illustrated in Figure \ref{fig:sampling}(a). In contrast, Fibonacci Lattices (FL) do not directly divide the polar angle and azimuth angle into a grid. Instead, they select sampling points on the sphere in a spiral pattern. As shown in Figure \ref{fig:sampling}(b), FL tends to achieve more uniform sampling, thereby improving the equivariance of EST. This is also consistent with the results observed in Table \ref{tab:equivariance}.

\paragraph{Enhanced Uniformity} Building on the FL sampling, we further propose a scattering optimization strategy similar to molecular dynamics. We define a distance-based interaction force between every two points: the greater the distance, the smaller the force. Then, we iterate two processes: 1. Simultaneously optimize the three-dimensional coordinates of all points through the interaction forces; 2. Re-project all points back onto the sphere. After multiple rounds of iteration, the spherical sampling points can become more uniform. It is worth noting that both the FL sampling and the dynamics simulation are conducted during the model initialization phase. Once we obtain the uniform sampling points, we save their results and directly read them in the model's feedforward process.

\paragraph{Self-conjugate Fourier Basis}
The orthogonality of the basis functions in equation \eqref{equ:fft} is defined by  $\int_{\Omega}Y^{(l,m)}(\vec{\mathbf{p}})Y^{(l',m')*}(\vec{\mathbf{p}})d\Omega=\delta_{ll'}\delta_{mm'}$, where $\delta_{ij}$ is the Kronecker delta. This work utilizes real-valued spherical harmonics, which possess the property of being self-conjugate: $Y^{(l,m)*}(\vec{\mathbf{p}}) = Y^{(l,m)} (\vec{\mathbf{p}})$.
Then we transform equation \eqref{equ:ifft} to a discrete inverse FT:
\begin{equation}
    x^{l,m} = \sum_{s=1}^{S} \lambda^{(l,m)}\mathbf{f}^{*}_{s} Y^{(l,m)*}(\vec{\mathbf{p}}_s) .
\end{equation}
To ensure orthogonality in the code implementation, we adopt the simple sum of squares rule to define the quadrature weights: $\lambda^{(l,m)} = \frac{1}{\sum_{s=1}^{S} |Y^{(l,m)}(\vec{\mathbf{p}}_{s})|^{2}}$.

\subsection{\revision{Approximation Argument for the Expressiveness Intuition}}\label{app:proof_expressiveness}
\revision{Claim \ref{theorem:expressiveness} is an unproven intuition rather than a formal function-class inclusion. This section explains why a sufficiently expressive spatial network can approximate CG-induced local products; Appendix \ref{app:controlled_ablation} supplies controlled empirical support and counterexamples showing additional cross-orientation capacity.}

Given two steerable vectors $\mathbf{u} \in \mathbb{V}_{0\to l_1}$ and $\mathbf{v} \in \mathbb{V}_{0\to l_2}$, and we define the CG tensor product result is $\mathbf{w} \in \mathbb{V}_{l_0}$, where $l_0 <= l_1 + l_2$. The spatial representations of $\mathbf{w}$ can be represented as:
\begin{equation}
\begin{aligned}
\label{equ:w}
    \sum_{l_0, m_0} \mathbf{w}^{(l_0,m_0)}Y^{(l_0,m_0)} = \sum_{l_0, m_0} ({\mathbf{u}^{(0\to l_1)}}^{T}\mathbf{C}^{(l_0,m_0)}_{(0\to l_1), (0 \to l_2)}\mathbf{v}^{(0\to l_2)})Y^{(l_0,m_0)},
\end{aligned}
\end{equation}
where $\mathbf{C}^{(l_0,m_0)}_{(0\to l_1), (0 \to l_2)}$ is a matrix including the whole CG coefficients corresponding to degree $l_0$ and order $m_0$. We temporarily ignore the input of the spherical harmonics. The spherical Transformer use the multiplication between spatial representations of $\mathbf{u}$ and $ \mathbf{v}$:
\begin{equation}
\begin{aligned}
\label{equ:mul_u_v}
    &\sum_{l_1,m_1}\mathbf{u}^{(l_1,m_1)}Y^{(l_1,m_1)} \sum_{l_2,m_2}\mathbf{v}^{(l_2,m_2)}Y^{(l_2,m_2)} \\
    =& \sum_{l_1,m_1,l_2,m_2}\mathbf{u}^{(l_1,m_1)}Y^{(l_1,m_1)}\mathbf{v}^{(l_2,m_2)}Y^{(l_2,m_2)} \\
    =&{\mathbf{u}^{(0\to l_1)}}^{T} ({\mathbf{Y}^{(0 \to l_1)}}^{T}\mathbf{Y}^{(0 \to l_2)})\mathbf{v}^{(0\to l_2)}
\end{aligned}
\end{equation}
Recall that CG coefficients are in fact the expansion coefficients of a product of two spherical harmonics in terms of a single spherical harmonic (see equation \eqref{equ:cg2sh1}):
\begin{equation}
\begin{aligned}
\label{equ:cg2sh}
    Y^{(l_1,m_1)} Y^{(l_2,m_2)}
  =& \sum_{l_0, m_0}
    \sqrt{\frac{(2 l_1 + 1) (2 l_2 + 1)}{4 \pi (2 l_0 + 1)}}
    \mathcal{C}_{\left(l_{1}, m_{1}\right)\left(l_{2}, m_{2}\right)}^{\left(0, 0\right)}
    \mathcal{C}_{\left(l_{1}, m_{1}\right)\left(l_{2}, m_{2}\right)}^{\left(l_{0}, m_{0}\right)}
    Y^{l_0, m_0}, \\
\end{aligned}
\end{equation}
equation \eqref{equ:mul_u_v} can be transformed to
\begin{equation}
\begin{aligned}
    &{\mathbf{u}^{(0\to l_1)}}^{T} ({\mathbf{Y}^{(0 \to l_1)}}^{T}\mathbf{Y}^{(0 \to l_2)})\mathbf{v}^{(0\to l_2)} \\
    =& \sum_{l_0, m_0}{\mathbf{u}^{(0\to l_1)}}^{T} \mathbf{H}^{(l_0,m_0)}_{(0\to l_1), (0 \to l_2)} \mathbf{v}^{(0\to l_2)}Y^{(l_0,m_0)},
\end{aligned}
\end{equation}
where $\mathbf{H}^{(l_0,m_0)}_{(l_1,m_1), (l_2,m_2)} = \sqrt{\frac{(2 l_1 + 1) (2 l_2 + 1)}{4 \pi (2 l_0 + 1)}}
    \mathcal{C}_{\left(l_{1}, m_{1}\right)\left(l_{2}, m_{2}\right)}^{\left(0, 0\right)}
    \mathcal{C}_{\left(l_{1}, m_{1}\right)\left(l_{2}, m_{2}\right)}^{\left(l_{0}, m_{0}\right)}$. Compared to $\mathbf{C}^{(l_0,m_0)}_{(0\to l_1), (0 \to l_2)}$, the term $\mathbf{H}^{(l_0,m_0)}_{(0\to l_1), (0 \to l_2)}$ introduces an additional constant term that can be approximated by linear layers and the FFNs. Therefore, equation \eqref{equ:mul_u_v} is equivalent to equation \eqref{equ:w}. On the other hand, equation \eqref{equ:mul_u_v} is inherently part of the Transformer architecture: when relationships between different orientations are masked and the $\mathbf{V}$ vectors are taken as constant vectors, the Transformer can naturally reduce to equation \eqref{equ:mul_u_v}.

Through the above process, we find that the spherical representation of the output from the CG tensor product is fundamentally a special case of EST. Furthermore, \emph{EST can capture dependencies between different orientations, which is particularly beneficial for approximating higher degree spherical harmonic representations. For instance, at a orientation $(\theta,\varphi)$, higher-degree spherical harmonics may involve contributions from other orientations such as $(2\theta,\varphi),(\theta,2\varphi),(2\theta,2\varphi)$ (see $Y^{(2)}$ in equation \eqref{equ:app_sh_real}). The Transformer in EST can directly combine these lower-degree terms in these orientations to approximate the higher-order terms at the orientation $(\theta,\varphi)$.}

We propose an alternative interpretation of the tensor product and spherical harmonics. The traditional tensor product can combine steerable vectors of any degree combination, and its functional form is bilinear. When we transform these steerable vectors to the spatial domain, the information from the different vectors, regardless of their degree, is projected onto a sphere. This projection allows a Transformer structure on a sphere to effectively capture the dependencies between the input steerable vectors. Furthermore, the Spherical Fourier Transform ($\mathcal{F}$) and its inverse ($\mathcal{F}^{-1}$) can be expressed as two matrix multiplications:
$ \mathcal{F}(\mathbf{x}) = W_{FT}\mathbf{x} = \mathbf{f}^{*} \text{and} \mathcal{F}^{-1}(\mathbf{f}^{*}) = W_{IFT}\mathbf{f}^{*}= \mathbf{x}$. Here, $W_{FT}$ and $W_{IFT}$ are matrices of size $S \times (L+1)^2$ and $(L+1)^2\times S$, respectively, which contain the Fourier basis. The vectors $\mathbf{x}$ and $\mathbf{f}^{*}$ are the steerable representation and spherical representation, with sizes $(L+1)^2 \times 1$ and $S \times 1$.

\paragraph{\revision{Controlled fitting evidence.}}\label{app:controlled_ablation}
\revision{We compare parameter-matched local FCTP (10,033 parameters), S2Activation (9,793), and EST (8,744) students on analytic local targets and on a nonlocal one-head attention target. Values below are test NRMSE over three seeds. The local rows support approximate correspondence between spectral CG products and pointwise spherical products at finite bandlimit. On the nonlocal target, explicit EST weights reproduce the teacher with maximum absolute error $1.19\times10^{-7}$, while the trained EST reaches $0.001872$ NRMSE versus $0.418440$ for FCTP and $0.435842$ for S2Activation. This is controlled evidence for the intuition in Claim \ref{theorem:expressiveness}, not a proof of unrestricted function-class inclusion.}
{
\begin{table}[htbp]
\centering
\caption{Controlled expressiveness fitting (test NRMSE, mean $\pm$ std.).}
\label{tab:controlled_expressiveness}
\small
\begin{tabular}{lccc}
\toprule
Target & FCTP & EST & S2Activation \\
\midrule
FCTP-local & $0.033470\pm0.000085$ & $0.021227\pm0.000748$ & $0.018035\pm0.000880$ \\
S2Activation-local & $0.034956\pm0.000022$ & $0.017678\pm0.000408$ & $0.015832\pm0.001622$ \\
EST-attention & $0.418440\pm0.000845$ & $\mathbf{0.001872\pm0.000088}$ & $0.435842\pm0.002330$ \\
\bottomrule
\end{tabular}
\end{table}}

\subsection{Equivariance of Relative Orientation Embedding}\label{app:proof_relativeembedding}
The equivariance of relative orientation embedding can be easily proven with the same way in equation \eqref{equ:proof_equivariance}, where $a\big(f^{Q}(\mathbf{R}\vec{\mathbf{p}}_{1}), f^{K}(\mathbf{R}\vec{\mathbf{p}}_{2})\big)$ is transformed to
\begin{equation}
    \frac{\exp\big(f^{Q}(\mathbf{R}\vec{\mathbf{p}}_{1})*f^{K}(\mathbf{R}\vec{\mathbf{p}}_{2}) + (\mathbf{R}\vec{\mathbf{p}}_{1})^T(\mathbf{R}\vec{\mathbf{p}}_{2})\big)}{\int_{\mathbf{R}\vec{\mathbf{p}}_{3}\in\Omega}\exp\big(f^{Q}(\mathbf{R}\vec{\mathbf{p}}_{1})*f^{K}(\mathbf{R}\vec{\mathbf{p}}_{3}) + (\mathbf{R}\vec{\mathbf{p}}_{1})^T(\mathbf{R}\vec{\mathbf{p}}_{3})\big)d\mathbf{R}\vec{\mathbf{p}}_{3}}.
\end{equation}
Due to the spherical symmetry, we can still eliminate the rotation $\mathbf{R}$ acting on $\mathbf{p}_{2}$ and $\mathbf{p}_{3}$. Therefore, after incorporating the Relative Orientation Embedding, EST retains its equivariance.

\begin{figure}
    \centering
    \includegraphics[width=0.8\linewidth]{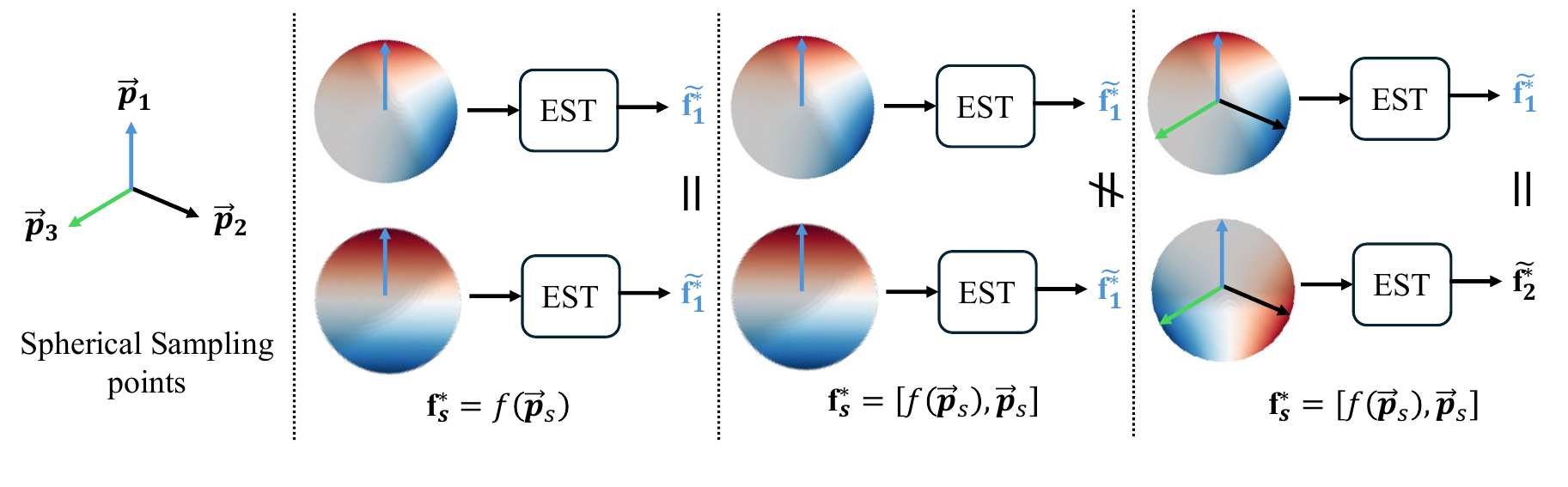}
    \caption{\textbf{An example of relative orientation embedding.} The three figures on the right illustrate distinct scenarios: the left and middle panels represent locally restructured spherical functions where the set of orientations has been reordered. Theoretically, the values in $\vec{\mathbf{p}}_1$ for these two spherical functions should differ after feeding EST; however, an EST without relative orientation embedding cannot distinguish between them. In contrast, the right panel illustrates globally rotated spherical functions. In this case, the spherical functions are identical under rotation, meaning the value in $\vec{\mathbf{p}}_1$ for the upper function should theoretically match the value in $\vec{\mathbf{p}}_2$ for the lower function after feeding EST.}
    \label{fig:position_embedding}
\end{figure}

We further elaborate on the contribution of the relative orientation embedding. In Figure \ref{fig:position_embedding}(left), EST without positional encoding cannot distinguish locally restructured spherical functions, but it can after using positional encoding (Figure \ref{fig:position_embedding}(middle)). In Figure \ref{fig:position_embedding} (right), the relative positional encoding ensures the overall equivariance of the spherical functions.

\subsection{Complexity Analysis}\label{app:complexity}
The complexity of the CG tensor product is related to the maximum degree $L$: $\mathcal{O}(L^6)$. In contrast, the complexity of the Transformer in EST is related to the spherical sampling number $ S $: $ \mathcal{O}(S^2) $ (self-attention). Based on the Nyquist sampling rate, the minimum number of samples on the spatial domain is at least $ (2L)^2 $, meaning the complexity of the Transformer in EST is $ \mathcal{O}(16L^4) $. The complexity of FFT and IFFT is $ 2 \cdot S \cdot L^2 \geq 8L^4 $. Therefore, the total complexity of EST is $ \mathcal{O}(24L^4) $. When $ L > 4 $, EST theoretically has lower computational cost. Moreover, in practical implementation, we also need to consider the impact of channel expansion on computational complexity. **Transformers can directly fuse multi-channel signals, allowing the channel dimension $ C $ in EST to be freely scaled, with complexity growing linearly with respect to $ C $: $ \mathcal{O}(24CL^4) $**. In contrast, the CG tensor product operation can only handle two single-channel irreducible representations at a time. When extending to multiple channels, if we consider pairwise interactions between all channels, the complexity increases to $ \mathcal{O}(C^2L^6) $. Some methods only consider interactions between a subset of channels (e.g., Equiformer DTP), reducing the complexity to $ \mathcal{O}(CL^6) $, but at the cost of reduced representational capacity. In summary, EST offers greater flexibility in channel expansion while maintaining lower computational complexity.

EST employs a standard fixed-length Transformer in its implementation, where the sequence length is $ S $, which enables it to benefit from various engineering optimizations. The results we presented in Figure \ref{fig:latency} already demonstrate that it can gain advantages from some hardware-friendly operations, such as 
\begin{verbatim}
import torch.nn.functional as F
out = F.scaled_dot_product_attention(q, k, v)
\end{verbatim}
For larger models, such benefits will be even more extensive, such as using Flash Attention and so on.

%% file: chapter/appendix/rebuttal_evidence.tex
\section{\revision{Controlled Ablations and Sampling Trade-offs}}\label{app:rebuttal_evidence}

\subsection{\revision{Controlled replacement in SCN}}\label{app:controlled_replacements}
\revision{We isolate cross-orientation attention in the original SCN implementation on OC20-S2EF-200K by replacing only its orientation-wise S2Activation with EST. We keep $L=3$, six interaction blocks, 256 hidden channels, 64 spherical channels, 20 neighbors, a \SI{6}{\angstrom} cutoff, and 60k training steps. EST increases parameters from 3.933M to 4.282M ($+8.9\%$).}
{
\begin{table}[htbp]
\centering
\caption{Controlled SCN replacement on OC20-S2EF-200K. Only S2Activation is replaced by EST.}
\label{tab:controlled_replacements}
\small
\resizebox{\linewidth}{!}{%
\begin{tabular}{lcccc}
\toprule
Model & Energy MAE $\downarrow$ & Force MAE $\downarrow$ & Force cosine $\uparrow$ & Force magnitude error $\downarrow$ \\
\midrule
SCN-original & \textbf{0.54533} & 0.05172 & 0.26900 & 0.08147 \\
SCN--EST & 0.56208 & \textbf{0.04931} & \textbf{0.29511} & \textbf{0.07597} \\
\bottomrule
\end{tabular}
}
\end{table}}
\revision{The SCN replacement reduces force MAE by $4.7\%$, improves force cosine by $9.7\%$, and reduces force-magnitude error by $6.7\%$, while energy MAE increases by $3.1\%$. Thus the controlled gain is concentrated on force modeling and cannot be attributed to a new spherical projection, which SCN already uses.}

\subsection{\revision{Sampling uniformity}}\label{app:sampling_uniformity}
\revision{We evaluate equal-size grids at $S\in\{16,36,64,100,144,196\}$ using nearest-neighbor angular-distance coefficient of variation (NN-CV), spherical-Voronoi area coefficient of variation (Area-CV), and spherical-cap discrepancy (Cap Error); lower is better. Table \ref{tab:sampling_uniformity} averages each metric across the six resolutions. Repulsion-refined FL ranks first overall at every resolution; plain FL ranks second.}
{
\begin{table}[htbp]
\centering
\caption{Sphere-sampling uniformity averaged over six equal-$S$ comparisons.}
\label{tab:sampling_uniformity}
\small
\begin{tabular}{lcccc}
\toprule
Sampling method & NN-CV $\downarrow$ & Area-CV $\downarrow$ & Cap Error $\downarrow$ & First place \\
\midrule
e3nn grid & 0.2934 & 0.4348 & 0.1511 & 0/6 \\
Fibonacci & 0.1091 & 0.0666 & 0.0691 & 0/6 \\
Fibonacci + repulsion & \textbf{0.0339} & \textbf{0.0337} & \textbf{0.0646} & \textbf{6/6} \\
\bottomrule
\end{tabular}
\end{table}}

\subsection{\revision{Resolution, accuracy, equivariance, and memory}}\label{app:resolution_tradeoff}
\revision{Table \ref{tab:resolution_tradeoff} uses a fixed QM9 configuration, batch size 4, and an RTX 4090. Accuracy and equivariance improve with $S$, but accuracy saturates rapidly while attention memory grows strongly. The Equiformer baseline uses 1,624 MiB. At the paper setting $S=128$, EST improves $\alpha$ MAE from 0.0460 to 0.0408 with equivariance error $1.05\times10^{-5}$ and 2,538 MiB peak memory; increasing to $S=512$ changes MAE by only 0.0003 but raises memory to 12,424 MiB. In the large EquiformerV2 experiment, adding EST raises same-batch single-GPU memory from 66,570 to 71,350 MiB; the smaller MPtrj global batch size was partly due to this overhead and limited GPU availability.}
{
\begin{table}[htbp]
\centering
\caption{Sampling-resolution trade-off on QM9.}
\label{tab:resolution_tradeoff}
\small
\begin{tabular}{lrrrr}
\toprule
Model & $\alpha$ MAE $\downarrow$ & Eq. error $\downarrow$ & Time/batch & Peak memory \\
\midrule
EqV baseline & 0.0460 & $8.482\times10^{-7}$ & 223 ms & 1,624 MiB \\
EST, $S=64$ & 0.0414 & $7.085\times10^{-4}$ & 246 ms & 1,944 MiB \\
EST, $S=128$ & 0.0408 & $1.050\times10^{-5}$ & 255 ms & 2,538 MiB \\
EST, $S=256$ & 0.0407 & $8.379\times10^{-6}$ & 268 ms & 4,790 MiB \\
EST, $S=512$ & 0.0405 & $3.501\times10^{-6}$ & 290 ms & 12,424 MiB \\
\bottomrule
\end{tabular}
\end{table}}
\revision{Practical guidance is therefore $S=\max\{64,(2L)^2\}$: the Nyquist term preserves the represented bandlimit and the lower bound avoids the large equivariance error observed at very coarse sampling. With abundant data, larger $S$ usually yields marginal accuracy gains; with scarce data or stricter equivariance requirements, $S$ can be increased subject to memory limits. Expressiveness is primarily adjusted through $L$ and the EST branch, whereas $S$ primarily controls discretization error and attention memory.}

\subsection{Physical validity for dynamics and structure optimization}\label{app:physical_validity}
We evaluate EST in two representative interatomic-potential settings: MPtrj with an independently predicted force head, and MD17 with conservative forces computed as $\mathbf F=-\nabla E$. For NVE molecular dynamics, initial velocities are sampled from a 300 K Maxwell--Boltzmann distribution with center-of-mass translation and rotation removed. Energy drift is the linear slope of total energy after discarding the first 20\% of each trajectory. On eight MD17 systems, EST reduces the mean absolute drift from 0.036 to 0.022 meV/ps/atom, and both models remain below 0.16 meV/ps/atom on every system.

Table \ref{tab:physical_validity} summarizes force-level and downstream physical checks. Under random rotations of approximately 100 MPtrj structures, both models retain cosine similarity above 0.9998 and force equivariance MAE below 0.6 meV/\AA. All perturbed structures converge under L-BFGS relaxation, and finite-difference Hessians introduce neither imaginary frequencies nor incorrect mode counts. These results support using EST for molecular dynamics, relaxation, and vibrational workflows when paired with an appropriate force construction.

\begin{table}[htbp]
\centering
\caption{Physical-validity tests. EqV1/2 denote the corresponding Equiformer baselines.}
\label{tab:physical_validity}
\small
\resizebox{\linewidth}{!}{%
\begin{tabular}{llcc}
\toprule
Setting & Metric & Baseline & Baseline + EST \\
\midrule
MD17, gradient force & Mean $|$NVE drift$|$ (meV/ps/atom; 8 systems) & 0.036 & \textbf{0.022} \\
MPtrj, direct force & Force equivariance MAE (meV/\AA; $\sim$100 structures) & \textbf{0.359} & 0.578 \\
MPtrj, direct force & Force cosine similarity & \textbf{0.99988} & 0.99987 \\
MPtrj, direct force & Jacobian asymmetry $\|J_{\mathrm{anti}}\|_F/\|J\|_F$ & \textbf{0.258} & 0.464 \\
MPtrj relaxation & Converged / average $F_{\max}$ (eV/\AA) & 16/16 / 0.041 & \textbf{16/16 / 0.038} \\
MD17 relaxation & Converged / average $F_{\max}$ (eV/\AA) & 8/8 / 0.014 & \textbf{8/8 / 0.011} \\
MPtrj vibrations & Imaginary frequencies / correct mode count & 0/16 / 16/16 & 0/16 / 16/16 \\
\bottomrule
\end{tabular}}
\end{table}

The direct-force MPtrj setting is not conservative for either model. EST increases the Jacobian-asymmetry metric from 0.258 to 0.464, and per-system MPtrj NVE drift is mixed rather than consistently improved. We therefore do not claim that EST repairs non-conservative force heads. Instead, the tests show that force equivariance, relaxation convergence, and vibrational stability remain practically intact, while conservative gradient forces on MD17 retain excellent long-term energy behavior.

%% file: chapter/appendix/supplementary_experiments.tex
\section{Details of Experiments and Supplementary Experiments}\label{app:exp}
\input{chapter/appendix/appendix_experiments/hyperparameters}

\begin{table}[htbp]
  \centering
  \caption{Hyper-parameters for the EST+GotenNet models on QM9 experiments.}
  \resizebox{\linewidth}{!}{
    \begin{tabular}{lll}
    \toprule[1.2pt]
    Hyper-parameters  & \multicolumn{2}{c}{EST (with GATA)} \\
    \midrule
    properties & $\alpha$, $\Delta\varepsilon$, $\varepsilon_{HOMO}$, $\varepsilon_{LUMO}$, $\mu$, $C_{v}$, $R^{2}$& $G$, $H$, $U$, $U_{0}$, U, ZPVE  \\
    Loss function  & MAE &  MSE \\
    Warmup steps & 0 & 10000 \\
    Weight decay & 0 & 0.01 \\
    Intermediate dimension and the number of steerable FFN & 512, 1 & 512, 4 \\
    Intermediate dimension and the number of spherical FFN & 512, 1 & 512, 4 \\
    Learning rate scheduling  & \multicolumn{2}{c}{linear warmup with reduce on plateau} \\
    Optimizer & \multicolumn{2}{c}{AdamW} \\
    Maximum learning rate & \multicolumn{2}{c}{$1 \times 10^{-4}$} \\
    Batch size & \multicolumn{2}{c}{32} \\
    Max training epochs & \multicolumn{2}{c}{1000} \\
    Dropout rate & \multicolumn{2}{c}{0.1} \\
    Number of RBFs & \multicolumn{2}{c}{64} \\
    Cutoff radius (Å) & \multicolumn{2}{c}{5} \\
    Lmax & \multicolumn{2}{c}{2} \\
    Number of layers & \multicolumn{2}{c}{4} \\
    Node dimension & \multicolumn{2}{c}{256} \\
    Edge dimension & \multicolumn{2}{c}{256} \\
    Number of attention heads & \multicolumn{2}{c}{8} \\
    Number of spherical point samples  & \multicolumn{2}{c}{64} \\
    \bottomrule[1.2pt]
    \end{tabular}}
  \label{tab:hyper_parameters_gotennet}%
\end{table}%

\input{chapter/appendix/appendix_experiments/ablation_modules}

%% file: chapter/appendix/appendix_experiments/hyperparameters.tex
\subsection{Implementation Details of Baselines}\label{app:hp_baseline}

\begin{table}[htbp]
\centering
\caption{Hyper-parameters for the EST model setting on OC20 S2EF and OC20 IS2RE experiments.
}\label{tab:hp_oc20}
\scalebox{0.65}{
\begin{tabular}{lll}
\toprule[1.2pt]
Hyper-parameters & S2EF-ALL & IS2RE  \\
\midrule[1.2pt]
Optimizer & AdamW & AdamW  \\
Learning rate scheduling & Cosine learning rate with linear warmup & Cosine learning rate with linear warmup \\
Warmup epochs & $0.01$ & $2$ \\
Maximum learning rate & $4 \times 10^{-4}$ & $2 \times 10^{-4}$ \\
Batch size & $256$ & $32$ \\
Number of epochs & $4$ & $20$ \\
Weight decay & $1 \times 10^{-3}$ & $1 \times 10 ^{-3}$ \\
Dropout rate & $0.1$ & $0.2$  \\
Energy coefficient $\lambda_{E}$ & $4,2$ & $1$  \\
Force coefficient $\lambda_{F}$ & $100$ & -  \\
Gradient clipping norm threshold & $100$ & $100$  \\
Model EMA decay & $0.999$ & $0.999$  \\
Cutoff radius (\AA) & $12$ & $5$ \\
Maximum number of neighbors & $20$ & $500$ \\
Number of radial bases & $600$ & $128$  \\
Dimension of hidden scalar features in radial functions & $128$ & $64$  \\

Maximum degree $L_{max}$ & $6$ & $1$  \\
Maximum order $M_{max}$ & $2$ & $1$\\
Number of Layers & $8$ & $6$ \\
Node embedding dimension  & $128$ & $(256, l=0),(128, l=1)$ \\

Intermediate dimension during the Fourier transform & $128$ & $256$ \\  
Intermediate dimension and the number of steerable FFN & $128, 6$ & $[(768, l=0), (384,l=1)], 5$ \\ 
Intermediate dimension and the number of spherical FFN & $512, 4$ & $512, 5$ \\ 
Number of spherical point samples & $156$ & $128$  \\

\bottomrule[1.2pt]
\end{tabular}
}
\vspace{2mm}
\label{appendix:tab:oc20_s2ef_hyperparameters}
\end{table}

\begin{table}[htbp]
\caption{Hyper-parameters for Mptrj dataset.}\label{tab:hyper_parameters_mptrj}
\centering
\scalebox{0.8}{
\begin{tabular}{lll}
\toprule[1.2pt]
Hyper-parameters & \multicolumn{1}{l}{EquiformerV2} &\multicolumn{1}{l}{EST} \\
\midrule
Optimizer & AdamW & AdamW \\
Learning rate scheduling & Cosine learning rate with & Cosine learning rate with \\
&linear warmup&linear warmup \\
Warmup epochs & $0.1$ & $0.1$ \\
Warmup factor & $0.2$ & $0.2$ \\
Maximum learning rate & $4 \times 10^{-4}$ & $4 \times 10^{-4}$ \\
Batch size & $512,256$ & $256$ \\
Number of epochs & $100$ & $100$ \\
Weight decay & $1\times10^{-3}$ & $1\times10^{-3}$ \\
Model EMA decay & $0.999$ & $0.999$ \\
Number of radial bases & \multicolumn{2}{c}{$512$ for Gaussian radial basis} \\
Cutoff radius (\AA) & $6$ & $6$ \\
$L_{max}$ & $4$ & $4$ \\
Number of layers & $8$  & $8$ \\
Node dimension & $128$ & $128$ \\
Energy loss coefficient & $5$ & $5$ \\
Force loss coefficient & $20$ & $20$ \\
Isotropic stress loss coefficient & $5$ & $5$ \\
Anisotropic stress loss coefficient & $5$ & $5$ \\
Intermediate dimension during the & $ $ & $512$ \\ 
Fourier transform & & \\
Intermediate dimension and the number of & $ $ & $128, 4$ \\ 
steerable FFN & & \\
Intermediate dimension and the number of & $ $ & $128, 4$ \\ 
spherical FFN & & \\
Number of spherical point samples & $ $ & $72$  \\
\bottomrule[1.2pt]
\end{tabular}
}
\end{table}

\begin{table}[htbp]
\caption{Hyper-parameters for QM9 dataset. }\label{tab:hp_qm9}
\centering
\scalebox{0.8}{
\begin{tabular}{lll}
\toprule[1.2pt]
Hyper-parameters & \multicolumn{1}{l}{Full-EST model (Figure \ref{fig:architecture})} &\multicolumn{1}{l}{EST+Equiformer} \\
\midrule
Optimizer & AdamW & AdamW \\
Learning rate scheduling & Cosine learning rate with & Cosine learning rate with \\
&linear warmup&linear warmup \\
Warmup steps & $5$ & $5$ \\
Maximum learning rate & $5 \times 10^{-4}$, $2 \times 10 ^{-4}$ & $5 \times 10^{-4}$, $1.5 \times 10^{-4}$ \\
Batch size & $128,64$ & $128,64$ \\
Max training epochs & $350,700$ & $300,600$ \\
Weight decay & $5\times10^{-3}, 0$ & $5\times10^{-3}, 0$ \\
Dropout rate & $0.0,0.2$ & $0.0,0.2$ \\
Number of radial bases & \multicolumn{2}{c}{$128$ for Gaussian radial basis $8$ for radial bessel basis} \\
Cutoff radius (\AA) & $5$ & $5$ \\
$L_{max}$ & $2$ & $2$ \\
Number of layers & $6$  & $6$ \\
Node dimension & $128$ & $(128,l=0),(64,l=1),(32,l=2)$ \\
Spherical harmonics embedding dimension & $(1,l=0),(1,l=1),(1,l=2)$ & $(1,l=0),(1,l=1),(1,l=2)$ \\
Intermediate dimension during the & $128$ & $128$ \\ 
Fourier transform & & \\
Intermediate dimension and the number of & $5, 128$ & $6, (768, l=0), (384,l=1)$ \\ 
steerable FFN & & \\
Intermediate dimension and the number of & $5, 512$ & $4, 512$ \\ 
spherical FFN & & \\
Number of spherical point samples & $200,128$ & $128$  \\
\bottomrule[1.2pt]
\end{tabular}
}
\end{table}

In the S2EF experiments, the results of baselines in Table~\ref{tab:oc20_s2ef} follow \cite{liao2024equiformerv}, where each model is trained in official configuration. Most of these configurations can be found in Fairchem repository, and we also follow its code framework to construct our OC20 experiments. In the IS2RE experiment, the results in Table~\ref{tab:IS2REDirect} of baselines follow \cite{liao2023equiformer} and \cite{zitnick2022spherical}. In the MPtrj experiments, the results are from \cite{batatia2023foundation,deng2023chgnet,barroso2024open}. In the QM9 experiments, the results in Table~\ref{tab:QM9} of baselines follow \cite{liao2024equiformerv} and \cite{aykentgotennet}.

\subsection{Implementation Details of EST Experiments}\label{app:hp_est}
\paragraph{S2EF and IS2RE} In our experiments on OC20, we adopt two hybrid models: S2EF combines the message module from EquiformerV2 with the update module of EST, while IS2RE combines the message module of Equiformer with the update module of EST. To ensure a fair comparison, all training configurations are aligned with those of the original EquiformerV2 and Equiformer. The hyperparameters specific to the EST architecture include the number of spherical sampling points, the number and dimension of experts in the steerable FFN, and the number of experts in the spherical FFN. Detailed configurations for all models are summarized in Table \ref{tab:hp_oc20}. The experiment on S2EF is conducted on 32 NVIDIA A100 GPUs and the experiment on IS2RE is conducted on 8 NVIDIA A100 GPUs. For comparison with SCN and eSCN, we report their test results due to the absence of publicly available validation results; previous work \citep{liao2024equiformerv} indicates that the validation and test sets have a similar distribution and the test results are usually slightly better than the validation results. 

\paragraph{MPtrj} In our experiments on MPtrj, we follow most of the configuration from prior work \citep{barroso2024open}. Due to limited computational resources, we reduced the batch size from 512 to 256. To ensure a fair comparison, we also reproduced the base model (EquiformerV2) under the same configuration in subsequent experiments. The settings of other parameters are shown in Table \ref{tab:hyper_parameters_mptrj}.

\paragraph{QM9} In the QM9 experiments, we design three EST variants: 1) a fully EST architecture (Table \ref{tab:QM9_full_est}) with both message and update blocks; 2) a hybrid model combining the message block of Equiformer and the update block of EST. We adopt similar configurations to base model as shown in Table \ref{tab:hp_qm9} and Table \ref{tab:hyper_parameters_gotennet}. Note that Equiformer employs two different configurations for the following properties: $\alpha, \Delta \varepsilon, \varepsilon_{\text{HOMO}}, \varepsilon_{\text{LUMO}}, \mu, C_v, R^2, \text{ZPVE}$, and $G, H, U, U_0$. We follow the same strategy in Table \ref{tab:hp_qm9}. GotenNet also has a similar division, and we follow its configuration.

%% file: chapter/appendix/appendix_experiments/ablation_modules.tex
\subsection{Supplementary GotenNet Experiments}\label{app:gotennet_experiments}

We offer further experiments on GotenNet. From Table \ref{tab:goten_QM9}, two principal observations emerge: first, EST improvements are inconsistent, primarily benefiting non-energy properties (e.g., $\alpha$, $\mu$, $\varepsilon_{HOMO}$) while struggling with core energy-related tasks ($G, H, U, U_0$). Second, the gains from EST diminish as GotenNet’s depth increases. In GotenNet-S, EST improves performance on six tasks, but yields no gains in the deeper GotenNet-L. This suggests that as deeper architectures potentially accumulate equivariance errors. While increasing FL sampling number could restore equivariance, it would incur substantial computational costs.

\begin{table}[htbp]
\caption{Results on QM9 dataset for various properties. $\dag$ denotes using different data partitions. }\label{tab:goten_QM9}
\resizebox{\linewidth}{!}{
\begin{tabular}{l|cccccccccccc}
\toprule
Task              & $\alpha$     & $\Delta \varepsilon$ & $\varepsilon_{HOMO}$ & $\varepsilon_{LUMO}$ & $\mu$ & $C_{v}$     & $G$ & $H$ & $R^{2}$  & $U$ & $U_{0}$ & $ZPVE$ \\
Units             & $bohr^{3}$     & meV                  & meV                  & meV                  & D     & cal$/$(mol K) & meV & meV & $bohr^{3}$ & meV & meV     & meV  \\ \midrule
GotenNet-B        &.032         &20.5         &15.2 &13.0 &.0072 &\textbf{.019} & 5.19 & 3.44 &.027 & 3.49 & 3.43 & 1.09 \\
+ EST    &.030         &31.6          &15.0 &13.1 &.0071 &\textbf{.019} & 5.13 & 3.60 &.028 & 3.58 & 3.55 & 1.17 \\
\midrule
GotenNet-L      &\textbf{.028}         & \textbf{19.8}          & \textbf{13.4} & \textbf{12.2} & \textbf{.0067} & \textbf{.019} & \textbf{4.98} & \textbf{3.30} & \textbf{.024} & \textbf{3.41} & \textbf{3.37} & \textbf{1.08} \\
+ EST    &\textbf{.028}         &30.8          & 13.7 & 12.8 &.0070 &\textbf{.019} & 5.26 & 3.41 &.027 & 3.50 & 3.53 & 1.18 \\
\bottomrule
\end{tabular}
}
\end{table}

\subsection{Supplementary Experiments}\label{app:ab_modules}

\paragraph{Theoretical experiments on expressiveness}

To rigorously evaluate the representational capacity of steerable architectures, we construct a synthetic dataset based on randomized tensor product operations. Let $\mathbf{x}_1, \mathbf{x}_2, \mathbf{x}_3 \in \mathbb{R}^3$ denote three independently sampled 3D vectors, corresponding to irreducible representations of degree $l=1$. The target output $\mathbf{y}$ is generated via a randomized Fully Connected Tensor Product (FCTP) network. To ensure comprehensive coverage of spherical harmonic degrees, we cascade two FCTP operations: $\mathbf{x} = \mathrm{FCTP}(\mathbf{x}_1, \mathbf{x}_2)$ yields an intermediate steerable feature spanning degrees $l \in \{0, 1, 2\}$, which is subsequently combined with $\mathbf{x}_3$ via $\mathbf{y} = \mathrm{FCTP}(\mathbf{x}, \mathbf{x}_3)$ to produce the final output spanning $l \in \{0, 1, 3\}$. A total of $N=1000$ samples are generated and partitioned into training and validation sets with a $9:1$ ratio.

We evaluate four distinct architectural variants:
\begin{enumerate}
    \item \textbf{FCTP-based model}: Mirrors the dataset generation pipeline but employs expanded channel dimensions for enhanced capacity.
    \item \textbf{EST}: A single-layer Equivariant Spherical Transformer.
    \item \textbf{S2Activation}: An ablated EST variant with the cross-orientational attention mechanism removed.
    \item \textbf{S2MLP}: A baseline that flattens all spherical sampling coordinates and channel dimensions into a monolithic vector, processed by a conventional multi-layer perceptron. This design discards explicit geometric structure.
\end{enumerate}

All models are optimized to regress $\mathbf{y}$ using the Mean Absolute Error (MAE) as the objective function. Each configuration is trained for a sufficient number of epochs to ensure convergence, and the procedure is repeated across $10$ independent random seeds to assess statistical stability. The resulting training and validation losses are reported in Table~\ref{tab:fctp_results}. Across all repetitions, the observed performance trends remain highly consistent: the FCTP and EST architectures achieve nearly identical losses, the S2MLP baseline attains marginally lower training loss but exhibits a validation gap, while the S2Activation variant demonstrates substantially elevated error.

\begin{table}[htbp]
    \centering
    \caption{Training and validation MAE on the FCTP-generated dataset. Results are averaged over 10 independent runs.}
    \label{tab:fctp_results}
    \begin{tabular}{lcccc}
        \toprule
        \textbf{Metric} & \textbf{FCTP} & \textbf{EST} & \textbf{S2Activation} & \textbf{S2MLP} \\
        \midrule
        Train loss & $2.58 \times 10^{-4}$ & $2.60 \times 10^{-4}$ & $3.31 \times 10^{-4}$ & $2.50 \times 10^{-4}$ \\
        Valid loss & $2.50 \times 10^{-4}$ & $2.52 \times 10^{-4}$ & $3.28 \times 10^{-4}$ & $2.68 \times 10^{-4}$ \\
        \bottomrule
    \end{tabular}
\end{table}

These findings substantiate our theoretical claim that the EST framework can effectively approximate arbitrary randomized FCTP operations. The S2MLP baseline, benefiting from an unconstrained architecture, attains the lowest training loss, reflecting its superior expressive capacity in unconstrained function approximation. However, its validation loss exceeds the training loss, indicating a tendency to overfit by memorizing dataset-specific patterns rather than learning an equivariant mapping. This behavior underscores a fundamental limitation of non-equivariant architectures in geometric learning tasks: they may exploit non-equivariant shortcuts to fit training data at the expense of generalization. Conversely, the S2Activation model consistently underperforms relative to both FCTP and EST across all experimental runs, evidencing its comparatively limited representational capacity.

The residual non-zero loss observed even in the FCTP-to-FCTP fitting task arises from the intrinsic stochasticity of the randomized tensor product generation process. For instance, the Clebsch--Gordan tensor product mapping two $l=1$ irreducible representations to an $l=1$ output is mathematically isomorphic to the vector cross product, which becomes degenerate when the input vectors are collinear or vanishing. To mitigate such multi-solution ambiguities and increase task complexity, we deliberately employ three distinct 3D vectors as inputs. Consequently, the generation process introduces irreducible stochastic components that cannot be perfectly reconstructed by a deterministic model, establishing a non-zero lower bound on the achievable loss.

To further probe the expressivity of the EST module, we conduct a complementary experiment using a dataset generated by a randomized EST layer. The corresponding performance metrics are summarized in Table~\ref{tab:est_results}. Notably, the unconstrained and semi-constrained architectures (EST, S2Activation, and S2MLP) achieve substantially lower losses, whereas the FCTP model exhibits a markedly higher error. This disparity strongly suggests that the EST module synthesizes representational features that transcend the functional space spanned by conventional tensor product operations.

\begin{table}[htbp]
    \centering
    \caption{Training and validation MAE on the EST-generated dataset.}
    \label{tab:est_results}
    \resizebox{\linewidth}{!}{%
    \begin{tabular}{lcccc}
        \toprule
        \textbf{Metric} & \textbf{FCTP} & \textbf{EST} & \textbf{S2Activation} & \textbf{S2MLP} \\
        \midrule
        Train loss & $7.06 \times 10^{-3}$ & $3.17 \times 10^{-6}$ & $8.04 \times 10^{-5}$ & $6.97 \times 10^{-5}$ \\
        Valid loss & $7.06 \times 10^{-3}$ & $3.21 \times 10^{-6}$ & $7.96 \times 10^{-5}$ & $9.13 \times 10^{-5}$ \\
        \bottomrule
    \end{tabular}
    }
\end{table}

All experimental scripts, data generation pipelines, and training routines are provided in \texttt{rebuttal/est\_vs\_tp.ipynb}. To facilitate independent verification of the architectural equivalence between EST and FCTP, trained model checkpoints can be serialized via \texttt{torch.save} for direct parameter inspection and cross-architecture comparison.

\begin{table}[htbp]
\centering
\scriptsize
\caption{The results of different expert configurations on OC20 IS2RE dataset.}
\label{tab:differentexpert_is2re}
\resizebox{1\linewidth}{!}{
\begin{tabular}{l|ccccc}
\hline
Model & energy (ID) MAE & energy (OOD Ads) MAE & energy (OOD Cat) MAE & energy (OOD Both) MAE & average \\ \hline
Equiformer (1 + 0)  & 504 & 688 & 521 & 630 & 586 \\
Equiformer + EST (5 + 5) & 501 & 652 & 502 & 578 & 558 \\
Equiformer + EST (0 + 1) & 507 & 664 & 513 & 593 & 569 \\
Equiformer + EST (1 + 1) & 502 & 657 & 504 & 588 & 563 \\ \hline
\end{tabular}
}
\end{table}

\begin{table}[htbp]
\centering
\scriptsize
\caption{Results of different expert configurations on QM9 dataset.}
\label{tab:differentexpert_qm9}
\begin{tabular}{l|ccc}
\hline
Model & $\alpha$ & $\varepsilon_{HOMO}$ & $U_0$ \\ \hline
Equiformer (1 + 0) & 0.046 & 15 & 6.59 \\
Equiformer + EST (4 + 6) & 0.041 & 14 & 5.64 \\
Equiformer + EST (0 + 1) & 0.045 & 15 & 6.85 \\
Equiformer + EST (1 + 1) & 0.042 & 14 & 5.73 \\
Equiformer + EST (10 + 0) & 0.043 & 15 & 6.47 \\
Equiformer + EST (8 + 0) & 0.044 & 15 & 6.41 \\
Equiformer + EST (4 + 0) & 0.050 & 15 & 6.68 \\
Equiformer + EST (2 + 0) & 0.046 & 15 & 6.74 \\ \hline
\end{tabular}
\end{table}
\begin{table}[htbp]
\centering
\scriptsize
\caption{Ablation studies for modules in EST.}\label{tab:ab_modules}
\begin{tabular}{ccccc|c}
\toprule
\multicolumn{5}{c}{Building blocks in EST} & $\alpha$ MAE  \\
\cmidrule(lr){1-5} \cmidrule(lr){6-6}
LayerNorm & SA & Steerable FFN & Spherical FFN & FL Sampling  & $bohr^{3}$ \\ \midrule
- & \checkmark & \checkmark & \checkmark & \checkmark & 0.046 \\
- & - & \checkmark & \checkmark & \checkmark & 0.044 \\
\checkmark & \checkmark & - & \checkmark & \checkmark & 0.045 \\
- & - & \checkmark & - & \checkmark & 0.043 \\
\checkmark & \checkmark & \checkmark & \checkmark & - & 0.053 \\
\checkmark & \checkmark & \checkmark & \checkmark & \checkmark & \textbf{0.041} \\
\bottomrule
\end{tabular}
\end{table}

\begin{table}[htbp]
\centering
\scriptsize
\caption{Results on Mptrj dataset. EquiformerV2 is replicated with the configuration in Table \ref{tab:hyper_parameters_mptrj}.}\label{tab:results_mptrj_replicate}
\begin{tabular}{l|cccc}
\toprule
model & energy (meV/atom) $\downarrow$ & forces (meV/\AA) $\downarrow$ & stress (meV/\AA$^3$)$\downarrow$ & forces cos $\uparrow$ \\
\midrule
EquiformerV2-S (Repetition) & 11.1 & 29.48 & 1.39 & 0.74 \\
EquiformerV2-S + EST & 10.80 & 28.85 & 1.36 & 0.74 \\
\bottomrule
\end{tabular}
\end{table}

\paragraph{Different expert configurations}
To further understand the contributions of EST and MoE, we conducted experiments on OC20 IS2RE and QM9 using various expert configurations. $(x+y)$ denotes $x$ steerable experts and $y$ spherical experts. As shown in Table \ref{tab:differentexpert_is2re} and Table \ref{tab:differentexpert_qm9}, we have several findings: 1.The update block using only spherical attention still achieves improvements. The minimal MoE configuration (1 steerable expert and 1 spherical attention expert) achieves results close to the multi-expert configuration.

\paragraph{Building Blocks} We conducted ablation studies to validate the effectiveness of individual building blocks within EST. Specically, in Table \ref{tab:ab_modules}, removing the steerable FFN means all experts are replaced by the spherical FFN, removing the spherical FFN means all experts are replaced by the steerable FFN and removing FL Sampling refers to using the Fourier transform from e3nn instead. We used the prediction of the $\alpha$ property on QM9 as the core task for evaluation. As shown in Table \ref{tab:ab_modules}, all components positively contribute to the overall performance. We draw several conclusions:
\begin{enumerate}
    \item The \textbf{SA module with layer normalization} effectively improves overall performance.
    \item \textbf{Mixing steerable and spherical experts} helps strike a balance between equivariance and expressive power, leading to better generalization performance.
    \item \textbf{FL Sampling is crucial for EST}. Disrupting equivariance without it significantly harms the results on QM9.
\end{enumerate}

\paragraph{Replication of EquiformerV2 on the MPtrj Benchmark} In our MPtrj experiments, we reproduced EquiformerV2 using the original configurations. As shown in Table \ref{tab:results_mptrj_replicate}, our replicated EquiformerV2 slightly outperforms the originally reported results. This improvement is likely attributable to a more conservative learning rate, which prevents the model from converging to suboptimal local minima. Furthermore, under these optimized baseline conditions, EST consistently yields significant performance gains, further validating its effectiveness in modeling complex molecular systems.

\begin{table}[htbp]
  \centering
  \scriptsize
  \caption{Efficiency benchmark of the proposed EST compared to DTP and FCTP baselines. Latency is reported in milliseconds (ms) under GPU acceleration across varying maximum degrees ($L$) and channels ($C$). Each data point represents the mean execution time across 50 independent trials to ensure statistical reliability.}
  \begin{tabular}{ccccccc}
    \toprule
    $L$ & Channels & EST Latency (ms) & DTP Latency (ms) & FCTP Latency (ms) & Speedup vs DTP & Speedup vs FCTP \\
    \midrule
    1 & 32 & 0.598 & 0.517 & 1.051 & 0.865 & 1.757 \\
    1 & 64 & 0.540 & 0.512 & 0.975 & 0.948 & 1.805 \\
    1 & 128 & 0.555 & 0.565 & 0.953 & 1.019 & 1.718 \\
    1 & 256 & 0.528 & 0.581 & 0.972 & 1.100 & 1.839 \\
    1 & 512 & 0.522 & 0.538 & 2.654 & 1.031 & 5.081 \\
    2 & 32 & 0.534 & 1.327 & 2.171 & 2.488 & 4.069 \\
    2 & 64 & 0.528 & 1.291 & 2.125 & 2.445 & 4.023 \\
    2 & 128 & 0.525 & 1.292 & 2.125 & 2.459 & 4.044 \\
    2 & 256 & 0.530 & 1.298 & 2.158 & 2.450 & 4.074 \\
    2 & 512 & 0.531 & 1.387 & 6.197 & 2.612 & 11.669 \\
    3 & 32 & 0.569 & 2.817 & 4.714 & 4.948 & 8.281 \\
    3 & 64 & 0.542 & 2.920 & 4.774 & 5.389 & 8.811 \\
    3 & 128 & 0.537 & 2.844 & 4.758 & 5.294 & 8.858 \\
    3 & 256 & 0.508 & 2.822 & 4.722 & 5.554 & 9.293 \\
    3 & 512 & 0.534 & 2.833 & 13.658 & 5.302 & 25.562 \\
    4 & 32 & 0.540 & 5.299 & 10.826 & 9.816 & 20.054 \\
    4 & 64 & 0.528 & 5.328 & 10.985 & 10.085 & 20.793 \\
    4 & 128 & 0.537 & 5.334 & 10.895 & 9.938 & 20.300 \\
    4 & 256 & 0.514 & 5.392 & 11.054 & 10.496 & 21.516 \\
    4 & 512 & 0.545 & 5.327 & 25.955 & 9.781 & 47.657 \\
    5 & 32 & 0.550 & 9.216 & 21.509 & 16.761 & 39.119 \\
    5 & 64 & 0.541 & 9.197 & 21.508 & 16.991 & 39.734 \\
    5 & 128 & 0.511 & 9.170 & 21.671 & 17.955 & 42.431 \\
    5 & 256 & 0.516 & 9.169 & 21.932 & 17.780 & 42.527 \\
    5 & 512 & 0.547 & 9.167 & 47.309 & 16.756 & 86.480 \\
    6 & 32 & 0.544 & 14.888 & 31.668 & 27.346 & 58.167 \\
    6 & 64 & 0.543 & 14.787 & 31.743 & 27.208 & 58.408 \\
    6 & 128 & 0.517 & 14.809 & 31.834 & 28.644 & 61.574 \\
    6 & 256 & 0.524 & 14.898 & 32.796 & 28.430 & 62.586 \\
    6 & 512 & 0.552 & 14.745 & 73.367 & 26.701 & \textbf{132.857} \\
    \bottomrule
    \end{tabular}%
  \label{tab:latency}%
\end{table}%

%% file: chapter/appendix/moe_details.tex
\section{\revision{Hybrid Branch and Architecture Details}}\label{app:moe_details}

\revision{The term ``hybrid experts'' denotes parallel function branches, not a large-language-model-style sparse MoE. Given an aggregated steerable message $\mathbf m$, the base model applies an exactly equivariant steerable FFN, while EST applies spherical attention and a spatial FFN after the Fourier transform. Invariant gate weights combine the branch outputs with the residual:}
{
\begin{equation}
\tilde{\mathbf m}=\mathbf m+\sum_{i=1}^{N_{\mathrm{steer}}}g_i\,\mathrm{FFN}^{\mathrm{steer}}_i(\mathbf m)+\sum_{j=1}^{N_{\mathrm{spatial}}}g_j\,\mathcal F^{-1}\!\left(\mathrm{FFN}^{\mathrm{spatial}}_j(\mathrm{EST}(\mathcal F(\mathbf m)))\right),
\end{equation}
}
\revision{where the gates are computed from the invariant $l=0$ channels. Our primary models use dense, small mixtures rather than top-$k$ routing. The purpose is to retain an exact-equivariant path alongside the expressive, finite-sampling EST path; it is not to scale parameter count through conditional computation. The ablations in Appendix \ref{app:ab_modules} show that adding same-type branches yields limited gains, whereas a minimal mixed $1+1$ configuration is consistently useful.}

\subsection{\revision{Integration into equivariant architectures}}\label{app:architecture}
\revision{EST can be inserted into either the edge-message or node-update block (Figure \ref{fig:architecture}). We use it mainly in update blocks because attention cost then scales with nodes rather than edges. A full-EST message block provides little additional QM9 benefit while increasing edge-dependent computation; Table \ref{tab:QM9_full_est} records this comparison.}
{
\begin{table}[htbp]
\caption{QM9 comparison of update-only and full EST variants.}\label{tab:QM9_full_est}
\resizebox{\linewidth}{!}{
\begin{tabular}{l|cccccccccccc}
\toprule
Task & $\alpha$ & $\Delta \varepsilon$ & $\varepsilon_{HOMO}$ & $\varepsilon_{LUMO}$ & $\mu$ & $C_v$ & $G$ & $H$ & $R^2$ & $U$ & $U_0$ & ZPVE \\
\midrule
Equiformer + EST update & .041 & 29 & 14 & 13 & .011 & .021 & 7.18 & 6.17 & .227 & 6.35 & 6.32 & 1.27 \\
Full EST & .042 & 28 & 13.4 & 12.4 & .0113 & .022 & 7.03 & 5.94 & .298 & 5.92 & 5.64 & 1.31 \\
\bottomrule
\end{tabular}}
\end{table}

\begin{figure}[t]
\begin{center}
\includegraphics[scale=0.32, trim=0cm 0cm 0cm 0cm,clip]{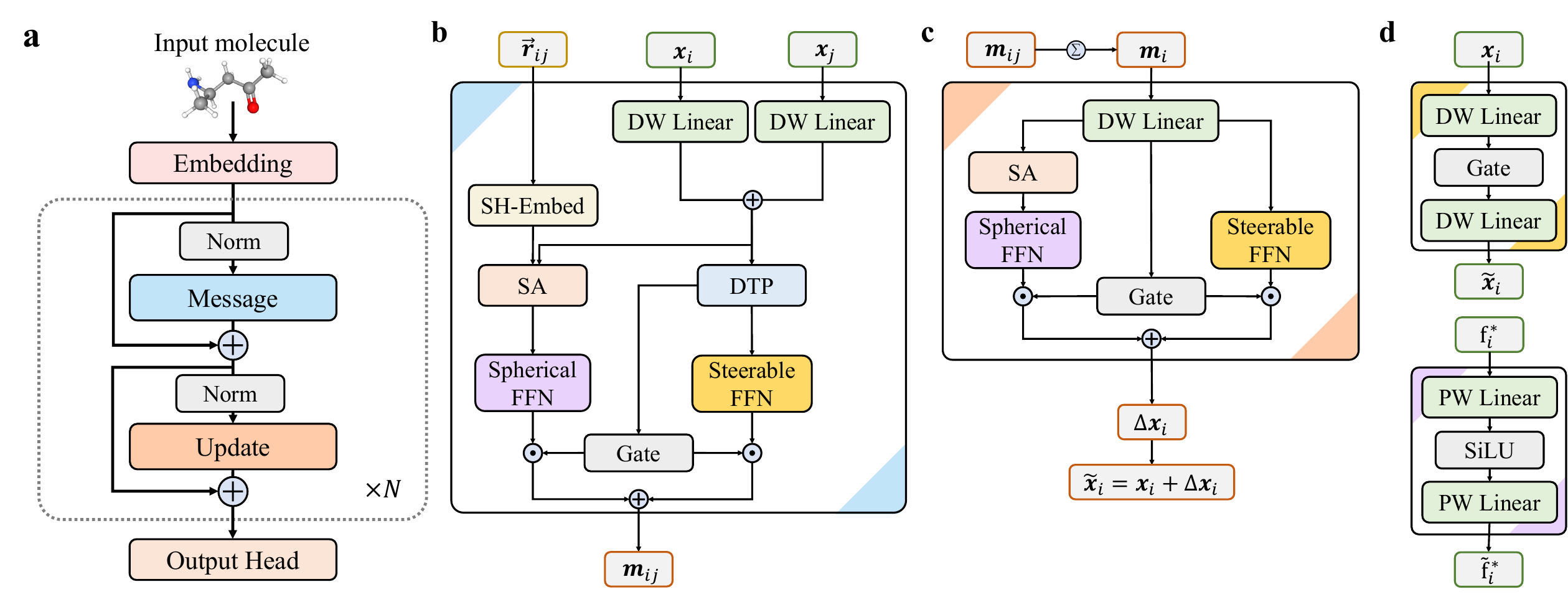}
\caption{\textbf{The architecture and building blocks of EST.} SH and DTP denote spherical harmonic embedding and depth-wise tensor product \citep{liao2023equiformer}. Fourier and inverse Fourier transforms are omitted for clarity. (a) Overall architecture. (b) Message block. (c) Update block. (d) Parallel steerable and spatial branches.}
\label{fig:architecture}
\end{center}
\end{figure}}